\documentclass{article}

\usepackage[final]{corl_2026} 
\usepackage{amsmath}
\usepackage{wrapfig}
\usepackage{algorithm}
\usepackage{algorithmic}
\usepackage{graphicx}
\usepackage{fontawesome}
\usepackage{multicol}
\usepackage[table]{xcolor}
\usepackage{csquotes}
\usepackage{booktabs}
\usepackage{multirow}
\usepackage{amssymb}
\usepackage{tcolorbox}
\tcbuselibrary{breakable}
\definecolor{citecolor}{HTML}{88A272}
\definecolor{linkcolor}{RGB}{192,0,0}
\hypersetup{
  pagebackref=true,
  breaklinks=true,
  colorlinks=true,
  citecolor=citecolor,
  linkcolor=linkcolor,
  bookmarks=false
}
\usepackage[capitalize]{cleveref}
\newtheorem{definition}{Definition}[section]

\newtheorem{remark}{Remark}[section]
\definecolor{spescolor}{RGB}{80,109,173}
\newcommand{\SPES}{\textcolor{spescolor}{\textbf{ReStruct}}}

\title{Inference-Time Robot Behavior Steering through Physically-Aware Reconfiguration of Task-Structure}

\author{Yiyuan Pan$^{1}$, Hanjiang Hu$^{1}$, Shangtao Li$^{1}$, Xusheng Luo$^{*,1}$, Changliu Liu$^{*,1}$\\
$^{*}$\textit{Co-advising}\\
Robotics Institute, Carnegie Mellon University, Pittsburgh, PA 15213, USA\\
\{yiyuanp, hanjianh, shangtal, xushengl, cliu6\}@andrew.cmu.edu \\
\vspace{-28pt}
}

\begin{document}
\maketitle


\begin{abstract}
A central challenge in deploying robot policies is \textit{inference-time policy steering}: redirecting a policy to satisfy natural-language preferences unseen in training, without retraining. Existing methods fail in two modes: end-to-end policies such as vision-language-action models (VLAs) require fine-tuning or expert-level guidance, while neuro-symbolic methods rely on predefined symbols whose edits can yield logically valid but physically infeasible plans. We propose \SPES, a zero-shot method built on a neural automaton policy that decomposes a visuomotor policy into a high-level state-machine skeleton and a low-level residual controller. A vision-language model (VLM) grounds the preference in the skeleton's learned symbols, refining them if needed, and compiles it into a temporal-logic automaton, whose synchronous product with the skeleton reconfigures the task structure. With the controller frozen, the skeleton's action priors are updated for physically-aware control. In simulation and real-world experiments, \SPES steers preferences from object-centric specifications to temporal-logic constraints, including novel combinations of demonstrated behaviors that require compositional generalization. Average preference-following success is 70.9\% in simulation and 92.5\% on a real robot, surpassing baselines like the prompt-steered vision-language-action model $\pi_{0.5}$.

\textbf{Webpage}: \href{https://pyywill.github.io/restruct-corl2026/}{\textcolor[RGB]{192,0,0}{https://pyywill.github.io/restruct-corl2026}}
\end{abstract}


\vspace{-5pt}
\begin{figure*}[htbp]
    \centering
    \includegraphics[width=\linewidth]{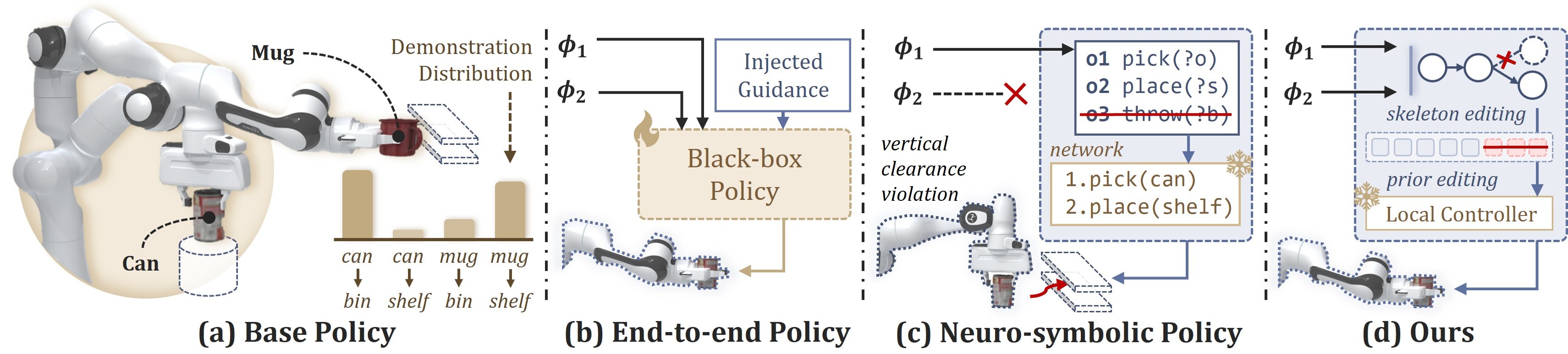}
    \vspace{-12pt}
    \caption{
    \textbf{Behavior Steering Mechanism for Different Models.} (\textbf{Preference}: $\phi_1$ ``\texttt{Put the can into the shelf}"; $\phi_2$ ``\texttt{Put the can to the higher place}") \textbf{(a)} The multi-task policy can place the can into the trash bin and the mug onto the shelf, respectively. \textbf{(b)} The end-to-end policy requires the injection of expert knowledge. \textbf{(c)} The neuro-symbolic policy either fails by generating a logically valid but physically infeasible plan ($\phi_1$), or fails due to its limited symbolic vocabulary ($\phi_2$). \textbf{(d)} Ours addresses policy steering through a three-stage process without requiring retraining.
    }
    \label{fig:teaser}
    \vspace{-10pt}
\end{figure*}

\section{Introduction}
\label{sec:intro}
Recent advances in robot learning have produced policies capable of diverse long-horizon manipulation \cite{mees2022calvin,zhou2023generalizable}. A central deployment challenge is adapting these policies to user preferences, constraints, or safety requirements not anticipated during training. We refer to it as \textit{inference-time behavior steering} \cite{wang2025inference,singhal2025general}, where policies must be steered at test time without retraining or additional data. Existing approaches generally fail in two modes: end-to-end methods, such as vision-language-action models \cite{black2025pi_,kim2024openvla} and diffusion policies \cite{chi2025diffusion}, adapt behavior either through fine-tuning or imposed guidance mechanisms, which are limited to preferences that are approximately differentiable and require expert knowledge. Neuro-symbolic methods \cite{bhuyan2024neuro,silver2022learning} support semantic-level preference specification through a symbolic layer, yet isolated symbolic modification may produce plans that are logically valid but physically infeasible, even for preferences expressible within the predefined symbol space.

At a conceptual level, both families essentially seek an intermediate policy-level representation (e.g., a soft Lagrangian in diffusion policies, predicates in neuro-symbolic methods) that translates semantic preferences into adjustments of the continuous policy distribution \cite{singhal2025general}. We argue that robust inference-time steering requires this representation to be \emph{physically-aware}: it should ground each preference in the subset of controller-executable behaviors that satisfy the preference. Existing methods fail this property in opposite ways (\cref{fig:teaser}). \textit{First}, end-to-end policies lack an explicit abstraction, so they must re-ground new preferences into the executable set through retraining or constraint injection \cite{nakamoto2024steering,haon2025mechanistic}. \textit{Second}, neuro-symbolic methods expose such an abstraction, but their symbolic edits cannot propagate to the controller-level networks; hence, even when the preference is expressible and the plan logically valid, execution may still stay mismatched and physically infeasible \cite{mao2025neuro}.

A recent method, Emergent Neural Automaton Policy (ENAP)~\cite{pan2026emergent}, offers a promising substrate for physically grounded steering. It decomposes a visuomotor policy into a low-level continuous controller and a high-level state-machine skeleton, whose symbolic abstractions can naturally serve as the intermediate representation between the semantic preference and the controller. Because the symbolic abstraction emerges automatically from demonstrations and is thus inherently aligned with the training data distribution, enabling in-domain preference editing. However, extending such representations to support policy steering under preferences remains an open problem.

To ensure physically-aware grounding to support efficient steering, we build on ENAP and introduce \SPES, a framework for adapting learned policies to user-specified preferences without retraining. Specifically, to satisfy a user-specified preference, \SPES\ first reconfigures the task structure of the base policy while keeping its low-level controller fixed. We use the automaton to represent the preference and incorporate it into the policy’s symbolic skeleton through a synchronous product. Since modifying the task structure alone can produce physically infeasible behaviors, \SPES\ subsequently reshapes the action priors provided the skeleton to filter out infeasible behaviors under the new specification, ensuring that the reconfigured structure remains executable by the controller. We validate \SPES\ from in-domain preferences to those that are physically in-domain yet semantically out-of-domain (OOD) through novel compositions. In summary, our contributions are:
\begin{itemize}
    \item We introduce \SPES, a steering framework that enables learned policies to follow high-level semantic preferences, ranging from object-centric specifications to temporal logic, offering everyday users a new perspective to reshape their robots through language, without expert-level knowledge or retraining.
    \item The framework provides interpretable and symbolically verifiable steering through its state-machine structure and prototype-based action priors, enabling preference satisfaction to be inspected at both the symbolic and control levels.
    \item The framework improves data efficiency and cross-task adaptation by reusing experience across diverse demonstrations. It enables target-task behavior to be synthesized by injecting preferences into demonstrations from related but distinct tasks.
    \item Extensive experiments show that \SPES\ improves preference following without degrading task success and surpasses VLA models by up to 25\% on both, indicating strategy adaptation rather than mere preference compliance.
\end{itemize}

\section{Steering via Bayesian Lens}
A Bayesian view factorizes the steering process into three terms, exposing what any steering method must control. 
Let $z$ denote the policy representation space, $\tau$ a trajectory admitted within $z$, and $\phi$ the user preference.

\textit{Firstly}, treating the base policy $p_{\mathcal D}(\tau)$ as prior, Bayes' rule gives $p(\tau\mid\phi)\propto p_{\mathcal D}(\tau)\,p(\phi\mid\tau)$ up to the constant $1/p(\phi)$, and marginalizing over $z$ yields $p(\phi\mid\tau)=\sum_z p(\phi\mid z,\tau)\,p_{\mathcal D}(z\mid\tau)$, with $p_{\mathcal D}(z\mid\tau)$ the demonstration-induced abstraction grounding. 
\textit{Secondly}, we assume conditional independence $\phi\perp\tau\mid z$: once $z$ is expressive enough to encode the preference, whether the preference is satisfied is determined by $z$ alone rather than by the particular trajectory realizing it, so $p(\phi\mid z,\tau)=p(\phi\mid z)$. \textit{Finally}, inverting $p(\phi\mid z)=p(z\mid\phi)\,p(\phi)/p_{\mathcal D}(z)$ gives the factorization
\begin{equation}
p(\tau\mid\phi)\;\propto\;p_{\mathcal D}(\tau)\sum\nolimits_z p_{\mathcal D}(z\mid\tau)\,\frac{p(z\mid\phi)}{p_{\mathcal D}(z)}.
\label{eq:exact}
\end{equation}
A uniform $p_{\mathcal D}(z)$ over its support (idealized equal-mass demonstration coverage) makes $1/p_{\mathcal D}(z)$ a constant absorbed into $\propto$:
\begin{equation}
\begin{aligned}
&p(\tau \mid \phi) \propto p_{\mathcal D}(\tau)\, p(\phi\mid\tau)\\
&\propto
\underbrace{p_{\mathcal D}(\tau)}_{\text{learned base policy}}
\sum_z
\underbrace{p_{\mathcal D}(z\mid\tau)}_{\text{abstraction grounding}}
\underbrace{p(z\mid\phi)}_{\text{preference grounding}}.
\end{aligned}
\label{eq:three-factor}
\end{equation}

Intuitively, each factor admits a direct reading: the base policy $p_{\mathcal D}(\tau)$ proposes trajectories, the abstraction grounding $p_{\mathcal D}(z\mid\tau)$ maps them to $z$-representation space, and the preference grounding up-weights the $\phi$-consistent $z$. Steering changes which symbolic paths are admissible under $\phi$, and must then re-connect the base policy's behaviors to the modified paths. Existing approaches cover only part of this factorization, which is the source of the two failure modes in \cref{sec:intro}: end-to-end methods provide only $p_{\mathcal D}(\tau)$, so $p_{\mathcal D}(z\mid\tau)\,p(z\mid\phi)$ must be injected via expert knowledge or retraining; neuro-symbolic methods edit the skeleton through $p(z\mid\phi)$ but omit $p_{\mathcal D}(z\mid\tau)$, so the edited skeleton may mismatch the low-level network.

Steering supplies a preference $\phi$, yet not every learned representation in $z$ is admitted by $\phi$: a representation encoding ``\textit{place the can into the trash bin}'' violates $\phi$ = ``\textit{place the can onto the shelf}''. At inference, we therefore need restructure $z$ into the feasible subspace $z'$ that $\phi$ admits. Restructuring alone, however, is insufficient, since $z'$ may still contain trajectories $\tau$ that are logically valid yet physically ungrounded, as in conventional neuro-symbolic methods. For instance, grasping the can with an upright end effector (\cref{fig:teaser}(c)) does not violate $\phi$, yet can severely disrupt subsequent decisions. Thus, a physically-aware refinement within $z'$ is also required to retain only the valid trajectories $\tau'$. $\tau'$ satisfies $p_{\mathcal D}(z'\mid\tau')=1$, which re-establishes grounding and brings controller-executable behavior back into the steered scope.

In the remaining context, we will provide further details on how we implemented the above-mentioned process: the policy representation $z$ takes the form of a state machine (PMM): we convert $\phi$ into a DFA and take its synchronous product with the skeleton ($p(z\mid\phi)$ term, \cref{sec:high-level}), and the low-level controller is held fixed when steering ($p_{\mathcal D}(\tau)$ term, \cref{sec:low-level}). Then, the per-edge action priors on the modified skeleton are recomputed ($p_{\mathcal D}(z\mid\tau)$ term, \cref{sec:bridge}), bridging the gap between the edited skeleton and the unchanged network to enable physically-aware control.

\section{Preliminaries}
\label{sec:preliminaries}
\paragraph{1) Emergent Neural Automaton Policy (ENAP).}
ENAP~\cite{pan2026emergent} decomposes a visuomotor policy into three parts: a high-level state machine skeleton, a low-level controller, and an action-prior bridge that transfers skeleton-level priors to the controller network. 

(i) \textit{High-level skeleton}: The skeleton is instantiated as a Probabilistic Mealy Machine (PMM):

\begin{definition}[Probabilistic Mealy Machine]\label{def:pmm}
A PMM is a tuple $\mathcal{M} = (Q, \Sigma, \Gamma, \delta, \lambda, q_0)$, where $Q$ is a finite set of states, $\Sigma$ a finite input alphabet, $\Gamma$ the output, $\delta(q' \mid q, \sigma), \sigma \in \Sigma$ the transition probability, $\lambda(\gamma \mid q, \sigma), \gamma\in\Gamma$ the output probability, and $q_0$ the initial state.
\end{definition}

In Definition~\ref{def:pmm}, states $q \in Q$ represent discrete task phases (e.g., \texttt{reach}, \texttt{grasp}), while $\Sigma$ denotes a discrete observation alphabet obtained by clustering encoded observations. Given an observation $o_t$, the encoder $\phi_\theta$ extracts a feature $z_t = \phi_\theta(o_t)$, which is then mapped to a cluster $\sigma_t \in \Sigma$. The continuous action space is denoted by $\Gamma = \mathcal{A} \subseteq \mathbb{R}^d$.

\begin{wrapfigure}{r}{0.27\textwidth}
    \vspace{-5mm}
    \centering
    \includegraphics[width=\linewidth]{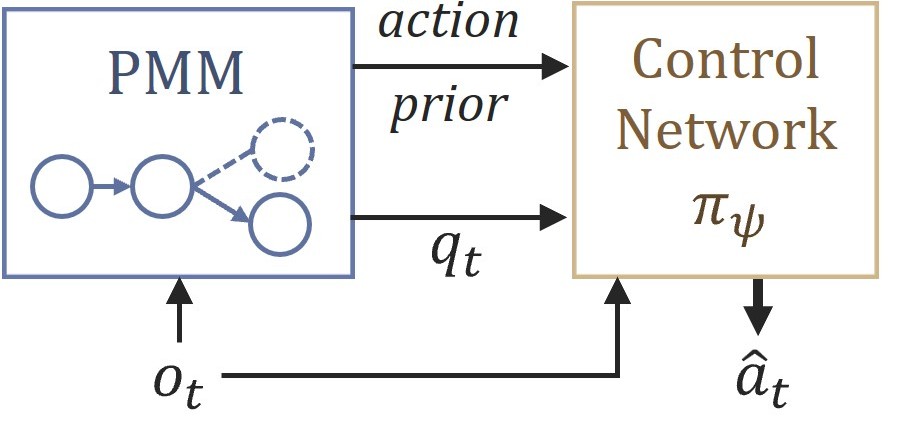}
    \vspace{-18pt}
    \caption{\textbf{ENAP diagram.}}
    \label{fig:enap}
    \vspace{-20pt}
\end{wrapfigure}
(ii) \textit{Action-prior Bridge}: Each edge $(q, \sigma) \to q'$ stores a fixed action priors $a^t_{\text{base}}$ drawn from demonstrations that exercise this transition. 

(iii) \textit{Low-level controller}: A residual policy network $\pi_\psi$ outputs a correction $\Delta a_t = \pi_\psi(q_t, o_t, a^t_{\text{base}})$. The final action is $\hat{a}_t = a^t_{\text{base}} + \Delta a_t$.

The inference process of ENAP follows a deterministic \textit{predict--act--update} cycle (\cref{fig:enap}). At time step $t$, given the current state $q_t$ and discretized observation input $\sigma_t$, the PMM first predicts a coarse action prior $a_{\text{base}}^{t}$. Guided by this prior, the low-level controller then generates fine-grained control actions. After execution, the PMM updates its internal state to $q_{t+1}$.

We build on ENAP because its symbolic abstraction emerges directly from demonstrations, providing natural alignment with the training data and the symbolic skeleton for in-domain steering. \SPES\ inherits ENAP's procedure for extracting the symbolic skeleton and observation alphabet, while reimplementing the bridge and low-level components to support physically-aware adaptation.

\paragraph{2) Linear Temporal Logic (LTL) Specifications.}
LTL extends propositional logic with temporal operators that describe how truth values evolve along a trajectory \cite{vardi2005automata}. Given a set of atomic propositions $\mathcal{AP}$, LTL formulas are built from $p \in \mathcal{AP}$, Boolean connectives, and the temporal operators $\bigcirc$ (next), $\Diamond$ (eventually), $\Box$ (always), and $\mathcal{U}$ (until). In our settings, these propositions correspond to clusters in the PMM's observation alphabet $\Sigma$, optionally refined by a VLM (e.g., splitting a generic ``\texttt{pick}'' cluster into ``\texttt{pick\_red\_cube}'' and ``\texttt{pick\_blue\_cube}''); a preference like ``\texttt{pick red cube first}'' is then expressible as $\Diamond(\texttt{\text{pick\_red\_cube}} \wedge \bigcirc \Diamond \texttt{\text{pick\_blue\_cube}})$. 
We consider a variant of LTL, namely Linear Temporal Logic over Finite Traces ($\mathrm{LTL}_f$), which is defined over finite-time robot trajectories \cite{de2013linear}. Each $\mathrm{LTL}_f$ formula can be translated into a Deterministic Finite Automaton (DFA) $\mathcal{A} = (S, s_0, \Sigma_\mathcal{A}, \delta_\mathcal{A}, F)$, where $\Sigma_\mathcal{A}$ denotes the input alphabet and $F \subseteq S$ is the set of accepting states. $s_0,\delta_\mathcal{A}$ denote the start state and transition function, respectively. The DFA accepts exactly the trajectories that satisfy the corresponding formula.

\begin{definition}[Synchronous Product of PMM \& DFA]
\label{def:product}
Given a PMM $\mathcal{M} = (Q, \Sigma, \Gamma, \delta, \lambda, q_0)$ and a DFA $\mathcal{A} = (S, s_0, \Sigma_\mathcal{A}, \delta_\mathcal{A}, F)$, their synchronous product is
\begin{equation}
\mathcal{P} = \mathcal{M} \otimes \mathcal{A} = (Q \times S, \, \Sigma_\otimes, \, \Gamma, \, \delta_\otimes, \, \lambda_\otimes, \, (q_0, s_0), \, F_\otimes),
\end{equation}
where the shared alphabet is $\Sigma_\otimes = \Sigma \cap \Sigma_\mathcal{A}$, the accepting set is $F_\otimes = Q \times F$, and the transition and output functions are
\begin{equation}
\delta_\otimes((q, s), \sigma) = \big(\delta(q, \sigma), \, \delta_\mathcal{A}(s, \sigma)\big), \qquad \lambda_\otimes((q, s), \sigma) = \lambda(q, \sigma).
\end{equation}
\end{definition}

We adopt $\mathrm{LTL}_f$ and DFAs for two reasons. \textit{First}, $\mathrm{LTL}_f$ provides a structured and expressive language for specifying preferences, ranging from object-centric descriptions to temporal-logic constraints. \textit{Second}, the synchronous product incorporates preferences into a state-machine representation in which every transition is traceable, making the steering verifiable and explainable.

\begin{figure*}
    \centering
    \includegraphics[width=\linewidth]{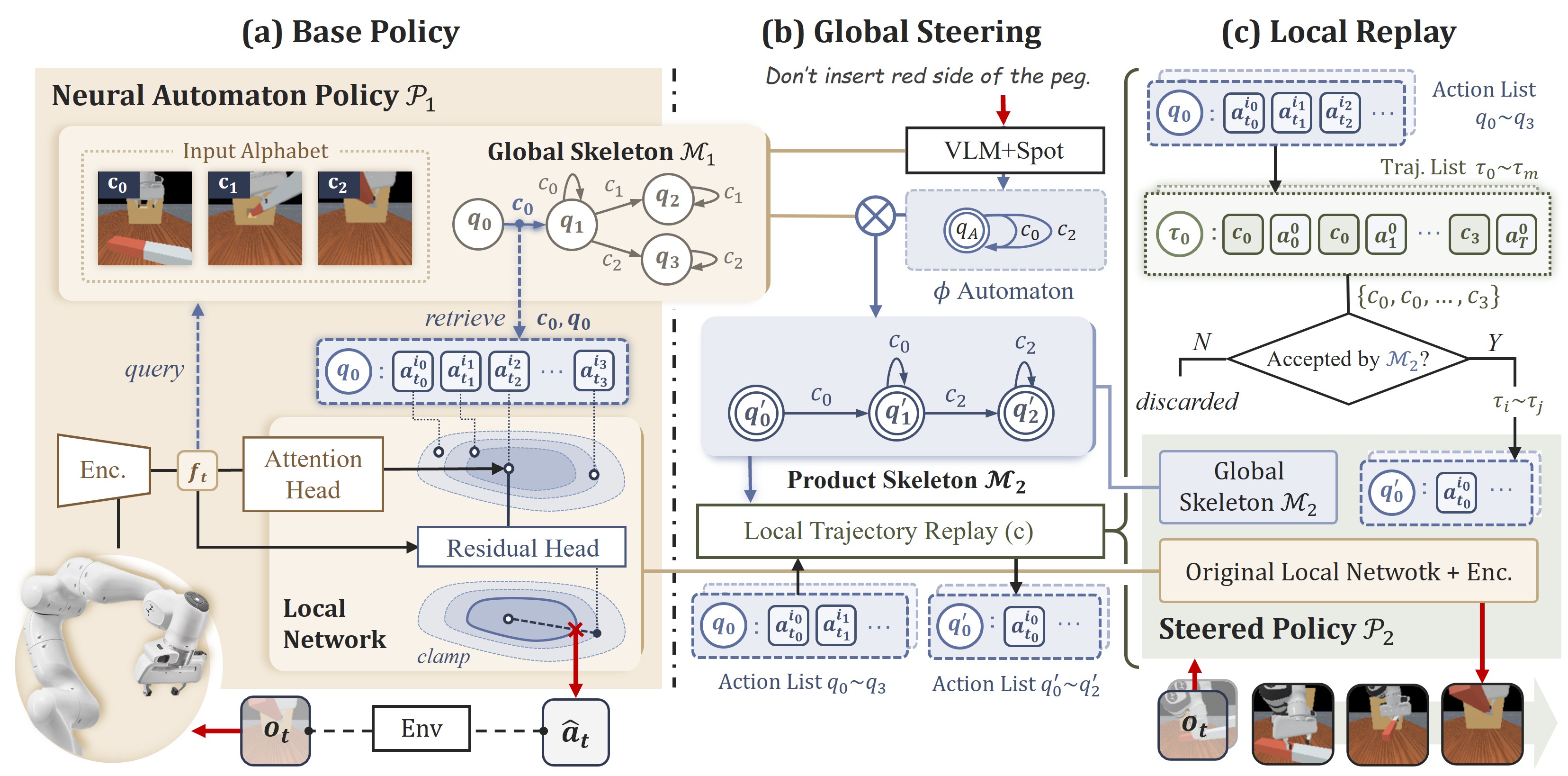}
    \vspace{-12pt}
    \caption{\textbf{Overview of \SPES} on \texttt{PegInsert} task. (\textbf{Left}) Base policy $\mathcal{P}_1$; (\textbf{Middle}) \SPES\ transforms preference into a DFA; its synchronous product with $\mathcal{M}_1$ yields the steered skeleton $\mathcal{M}_2$. (\textbf{Right}) Trajectories are repopulated, producing $\mathcal{P}_2$, which inherits $\mathcal{P}_1$'s frozen low-level. \textcolor[RGB]{92,111,158}{\textbf{Blue}} / \textbf{Black} / \textcolor[RGB]{192,0,0}{\textbf{Red}} lines are high-level skeleton / low-level controller / user-side flows.
}
    \label{fig:overview}
    \vspace{-10pt}
\end{figure*}

\section{Methodology}
\paragraph{Problem Formulation.}
We consider a Partially Observable Markov Decision Process (POMDP) \cite{spaan2012partially,kurniawati2022partially} with observation space $\mathcal{O}$ and continuous action space $\mathcal{A}$, in which a trajectory takes the form $\tau = \{(o_t, a_t)\}_{t=0}^T$. We are given a base policy $\pi_0$ trained on a demonstration set $\mathcal{D} = \{\tau_i\}_{i=1}^N$. At deployment, the user provides a natural-language preference $\phi$. The goal is to produce an adapted policy $\pi^\phi$ that maximizes $\Pr_{\tau \sim \pi^\phi}[\tau \models \phi]$ while preserving the task performance without retraining. Equivalently, $\pi^\phi$ targets the posterior $p(\tau \mid \phi) \propto p_\mathcal{D}(\tau) \cdot p(\phi \mid \tau)$.

\subsection{High-Level: Symbolic Skeleton Reconfiguration}
\label{sec:high-level}
\begin{wrapfigure}{r}{0.471\textwidth}
    \vspace{-3mm}
    \centering
    \includegraphics[width=0.98\linewidth]{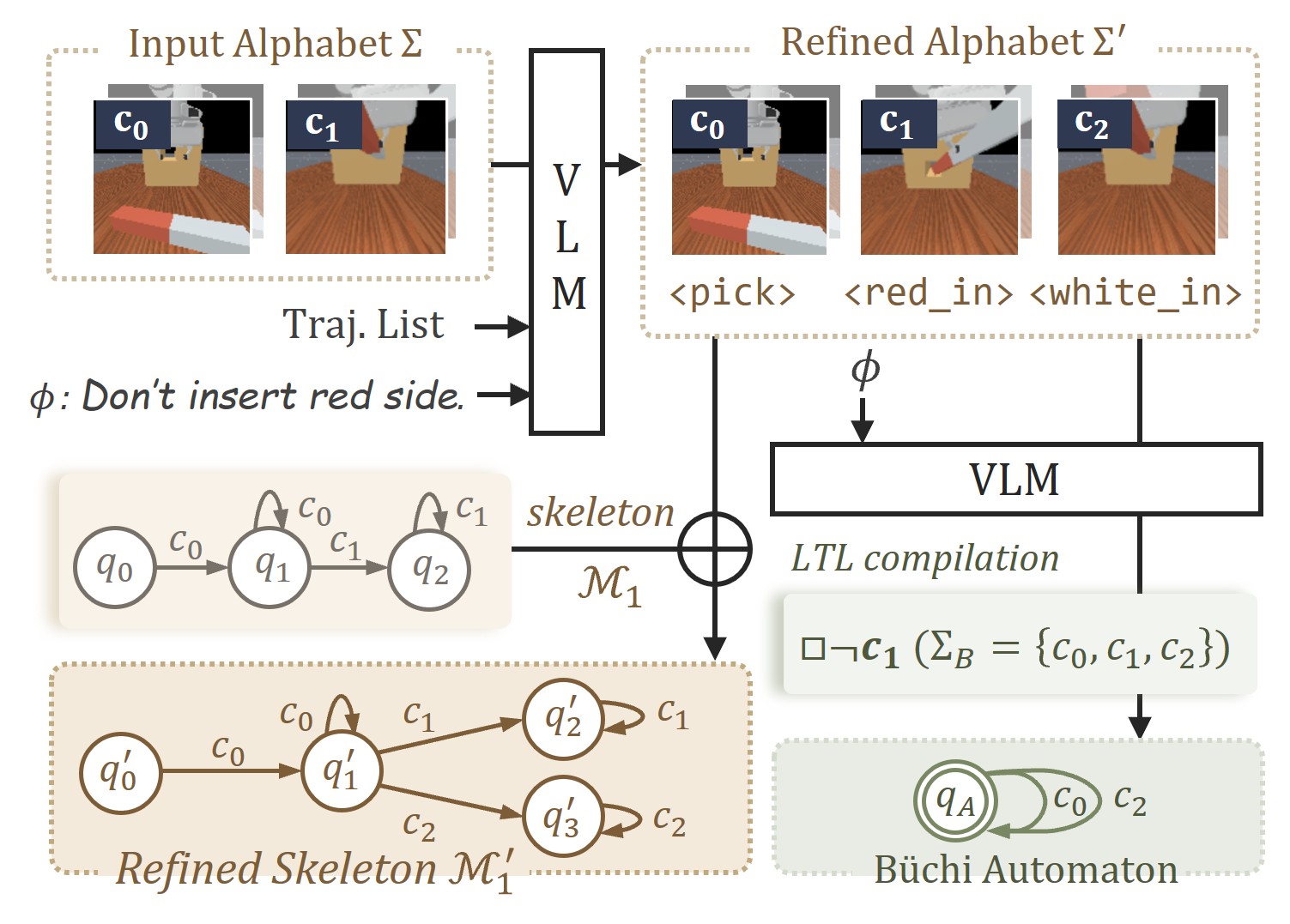}
    \vspace{-5pt}
    \caption{\textbf{VLM-based Preference Grounding.} The VLM refines the alphabet and skeleton $\mathcal{M}_1$, then generates preference DFA.}
    \label{fig:vlm}
    \vspace{-12pt}
\end{wrapfigure}
This subsection instantiates the ${p(z|\phi)}$ term of \cref{eq:three-factor}: Given a user preference $\phi$, we construct a steered skeleton $\mathcal{M}_2$ by reconfiguring the base skeleton $\mathcal{M}_1$, such that the resulting skeleton can generate only preference-satisfying plans. The construction proceeds in two stages: (i) DFA generation from the preference $\phi$; (ii) synchronous product between the resulting DFA and the base skeleton $\mathcal{M}_1$. The structure of the base skeleton and its clustering alphabet may be refined.

A VLM first receives three inputs: the cluster alphabet $\Sigma$ from the base skeleton $\mathcal{M}_1$, the natural-language preference $\phi$, and $\mathcal{M}_1$ itself. It returns a DFA $\mathcal{A}$ together with a (possibly refined) skeleton $\mathcal{M}_1’$. Internally, the VLM first determines whether the clustering alphabet $\Sigma$ is expressive enough to encode the preference $\phi$: 
(i) If $\Sigma$ is sufficient, the VLM translates $\phi$ into an $\mathrm{LTL}_f$ formula~\cite{xu2025nl2hltl2plan},
compiles it into a DFA $\mathcal{A}$ using toolbox Spot~\cite{duret2016spot}, and leaves the skeleton unchanged ($\mathcal{M}_1’=\mathcal{M}_1$).
(ii) If $\phi$ requires finer semantic granularity, the VLM refines the relevant clusters (e.g., \texttt{pick} $\rightarrow$ {\texttt{pick\_red}, \texttt{pick\_blue}}) to obtain an expanded alphabet $\Sigma'$. We then generate $\mathcal{A}$ over $\Sigma’$, and refine $\mathcal{M}_1$ by replacing each affected edge $(q,\sigma,q') \in \mathcal{M}_1$ with the set
$\{(q, \sigma_k, q') : \sigma_k \in \mathrm{split}(\sigma)\}$.
During the process, only the topological structure is modified; the action priors remain unchanged. The procedure is shown in \cref{fig:vlm}.

We then form the steered skeleton $\mathcal{M}_2 \;=\; \mathcal{A} \otimes \mathcal{M}^{'}_1$ following the \cref{def:product}. Intuitively, the product takes the intersection of two logics: $\mathcal{M}^{'}_1$ captures the traces induced by the demonstrations, $\mathcal{A}$ admits the traces satisfying $\phi$. The resulting $\mathcal{M}_2$ serves as the structural target against which action priors are re-computed in the next subsection.

\subsection{Action-Prior Bridge: Trajectory Replay}
\label{sec:bridge}
In the previous discussion, we showed that refining the $p_{\mathcal{D}}(z|\tau)$ term of the steered skeleton is vital: the traditional neuro-symbolic method fails in steering because its prior is anchored to the demonstration that may contain preference-violating behaviors. In our paper, the high-level product (\cref{sec:high-level}) yields $\mathcal{M}_2$ that admits solely preference-consistent traces, and we need to recompute the action priors for the skeleton to guide physically-aware generation of preference-aligned behavior.

Not only for neuro-symbolic methods, most visuomotor policies learn a direct observation-to-action map, compressing the separable structure in the data into an entangled representation. For example (see \cref{fig:teaser} (c)), when executing $\texttt{pick(can)} \to \texttt{place(shelf)}$, such a single-mode prior grasps the can from above and incurs a vertical-clearance violation that blocks shelf insertion. ENAP's per-edge fixed action has the same failure mode. Thus, we redesign the action-prior bridge for ENAP: \SPES\ stores, for each edge $(q, \sigma) \to q'$, the raw list of the action that performs such transitions in the demonstration, to provide diverse priors. Empirically, during skeleton construction, we randomly retain $1/50\sim 1/10$ of the demonstrations as representative trajectories to make it lightweight.

Recognizing the steered skeleton $\mathcal{M}_2$ as an acceptor, we replay each representative trajectory $\tau_i=\{c_t^i,a_t^i\}_{t=0}^{T_i}$, following from the initial node $q_0{'}=(q_0,s_0)$ and the transitions its clusters induce. A trajectory is \emph{rejected} and discarded if it reaches a state with no valid transition or terminates outside the accepting set $F_\otimes$ of the synchronous product $\mathcal{M}_2$; otherwise it is \emph{accepted}, and its actions are written into the action lists of the edges it traverses.
Edges that no accepted trajectory traverses are pruned. The populated action list becomes the action-priors' candidate pool from which the low-level controller selects at inference.

\begin{figure*}[t]
    \centering
    \includegraphics[width=\linewidth]{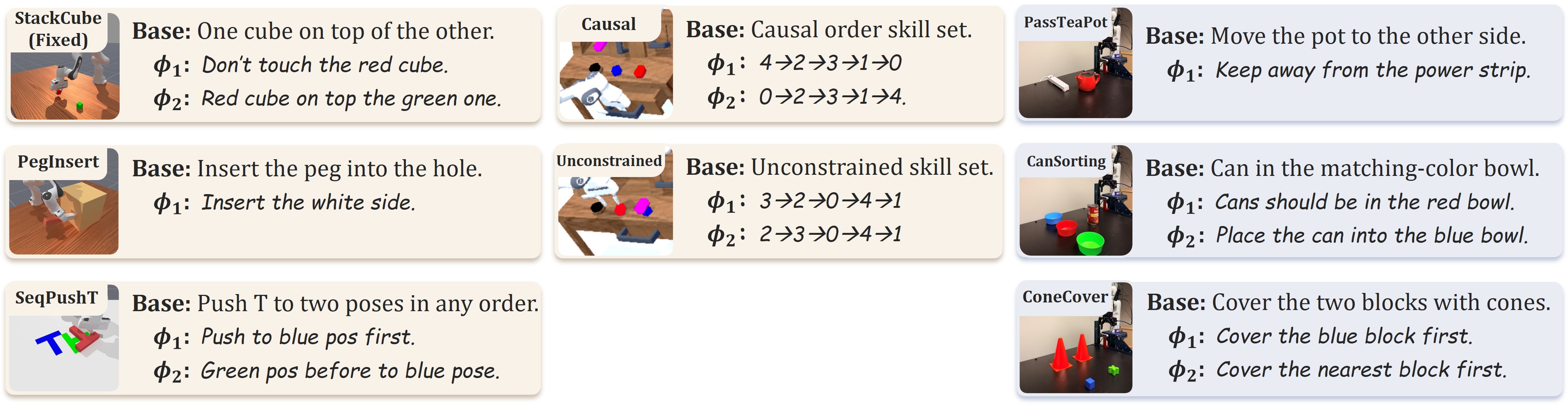}
    \vspace{-12pt}
    \caption{\textbf{Task Suite}. \texttt{Causal} skill 0$\sim$4: \textit{rotate\_block, close\_drawer, open\_drawer, push\_right, push\_left}; \texttt{Unconstrained} skill 0$\sim$4: \textit{turn\_on\_led, turn\_on\_bulb, rotate\_block, push\_left, move\_slider}. \textbf{Challenges}: \texttt{PassTeaPot} demands a trajectory-level preference, \texttt{CanSorting}'s $\phi$ requires compositional generalization, e.g., a new combination of red can with the blue bowl. \texttt{ConeCover} tests spatial reasoning with task memory, since a placed cone shouldn't be re-picked.
    \texttt{Causal} and \texttt{Unconstrained} require combine atomic skills to form long-horizon execution.
    }
    \label{fig:exp}
    \vspace{-1.8pt}
\end{figure*}

\subsection{Low-Level: Action Generation}
\label{sec:low-level}
The controller is frozen throughout steering. Its design must therefore stay reliable when the prior inherited from the high-level skeleton it consumes changes. Thus, we also redesign the structure of the low-level controller for ENAP, and keep it frozen after network training. Specifically, the controller process the observation $o_t$ and the active edge's action list $a_t^{\text{list}}$ through following process:
\begin{equation}
\begin{aligned}
&a_t^{\text{prt}} = \text{\textbf{SoftAttn}}\big(\phi_\theta(o_t),\, a_t^i\big), \quad \forall\, a_t^i \in a_t^{\text{list}}, \\
&a_t^{\text{raw}} = a_t^{\text{prt}} + \text{\textbf{MLP}}\big(\phi_\theta(o_t) - c_t,\, a_t^{\text{prt}}\big),   \\
&\hat{a}_t = \text{\textbf{Clamp}}\big(a_t^{\text{raw}},\, a_t^{\text{list}}\big).
\end{aligned}
\label{eq:lowlevel}
\end{equation}
Specifically, attention maps $\phi_\theta(o_t)$ to a prototype $a_t^{\text{prt}}$; a residual MLP combines it with the within-cluster offset $\phi_\theta(o_t) - c_t$ ($c_t$: centroid of the active cluster $\sigma_t$) to produce $a_t^{\text{raw}}$, which is clamped to the range of $a_t^{\text{list}}$ to generate the executable $\hat{a}_t$.
The encoder, attention module, and MLP are jointly trained via behavior cloning (ENAP's training algorithm) and frozen during steering. Two reasons motivate this attention design over ENAP's fixed prior (see action-prior bridge, \cref{sec:preliminaries}). \textit{First}, it makes the MLP robust to steering-induced prior changes: the fixed prior would expose the MLP to a limited set of priors during training, so when steering rewrites the structure, the new prior could be OOD. \textit{Second}, prototype priors give the controller finer adaptability and flexibility.

\begin{remark}[Steerable Scope]
\label{rem:scope}
\SPES\ considers: (i) \textbf{In-domain}: the preferred trajectory is demonstrated, whether or not semantic clustering exposes it; (ii) \textbf{Compositional OOD}: the trajectory itself is unseen, yet its behaviors are each demonstrated, semantically compositional---novel combination of previously seen behaviors (e.g., \emph{\texttt{CanSorting}} $\phi_1$ in \cref{fig:exp}) \cite{wang2023programmatically}. 
\end{remark}

\section{Experiments}
In this section, we present extensive experimental results to
address the following questions:
\textbf{(Q1)}: Can \SPES\ improve preference adherence without degrading task success? \textbf{(Q2)}: Can \SPES\ assemble ordered sequences from unordered data to complete long-horizon tasks? \textbf{(Q3)}: Can \SPES\ exhibit the steering advantages of a state machine over other symbolic structures?

\begin{table}[t]
\centering
\caption{\textbf{Comparison on Complex Manipulation.} We report mean values over 8 seeds 64 trials.}
\label{tab:maniskill}
\renewcommand{\arraystretch}{1.1}
\resizebox{\linewidth}{!}{%
\begin{tabular}{l @{\hspace{2.5mm}} ccccc c@{\hspace{2.5mm}} ccc @{\hspace{2.5mm}} ccccc}
\toprule
\multirow{4}{*}{\textbf{Method}}
& \multicolumn{5}{c}{\texttt{StackCube (Fixed)}}
& \multicolumn{3}{c}{\texttt{PegInsert}}
& \multicolumn{5}{c}{\texttt{SeqPushT}} \\
\cmidrule(lr){2-6}\cmidrule(lr){7-9}\cmidrule(lr){10-14}
& \multicolumn{1}{c}{Base} & \multicolumn{2}{c}{$\phi_1$} & \multicolumn{2}{c}{$\phi_2$}
& \multicolumn{1}{c}{Base} & \multicolumn{2}{c}{$\phi_1$}
& \multicolumn{1}{c}{Base} & \multicolumn{2}{c}{$\phi_1$} & \multicolumn{2}{c}{$\phi_2$} \\
\cmidrule(lr){2-2}\cmidrule(lr){3-4}\cmidrule(lr){5-6}\cmidrule(lr){7-7}\cmidrule(lr){8-9}\cmidrule(lr){10-10}\cmidrule(lr){11-12}\cmidrule(lr){13-14}
& \textbf{SR} & \textbf{SR} & \textbf{SRP} & \textbf{SR} & \textbf{SRP}
& \textbf{SR} & \textbf{SR} & \textbf{SRP}
& \textbf{SR} & \textbf{SR} & \textbf{SRP} & \textbf{SR} & \textbf{SRP} \\
\midrule
\multicolumn{14}{l}{\cellcolor{gray!15}\textit{- Behavior Cloning (Reject Filtering)}} \\
Transformer \cite{vaswani2017attention}
& $89.1$ & $29.7$ & $29.7$ & $59.4$ & $59.4$
& $90.6$ & $60.9$ & $60.9$
& $37.5$ & $21.9$ & $21.9$ & $15.6$ & $15.6$ \\
GMM \cite{reynolds2009gaussian}
& $87.5$ & $42.2$ & $42.2$ & $45.3$ & $45.3$
& $85.9$ & $53.1$ & $53.1$
& $\mathbf{53.1}$ & $34.4$ & $34.4$ & $21.9$ & $21.9$ \\
\multicolumn{14}{l}{\cellcolor{gray!15}\textit{- Diffusion Policy (Soft Lagrangian)}} \\
DP \cite{chi2025diffusion}
& $93.8$ & $87.5$ & $68.8$ & $\mathbf{81.2}$ & $42.2$
& $57.8$ & $42.2$ & $20.3$
& $40.6$ & $28.1$ & $18.8$ & $31.2$ & $18.8$ \\
\multicolumn{14}{l}{\cellcolor{gray!15}\textit{- VLA Models (Prompt Injection)}} \\
OpenVLA \cite{kim2024openvla}
& $81.2$ & $67.2$ & $21.9$ & $57.8$ & $28.1$
& $68.8$ & $54.7$ & $29.7$
& $18.8$ & $20.3$ & $12.5$ & $7.8$ & $3.1$ \\
$\pi_{0.5}$ \cite{black2025pi_}
& $92.2$ & $85.9$ & $37.5$ & $73.4$ & $48.4$
& $85.9$ & $75.0$ & $50.0$
& $28.1$ & $31.2$ & $26.6$ & $12.5$ & $6.2$ \\
\midrule
\SPES
& $\mathbf{95.3}$ & $\mathbf{98.4}$ & $\mathbf{98.4}$ & $76.6$ & $\mathbf{76.6}$
& $\mathbf{98.4}$ & $\mathbf{95.3}$ & $\mathbf{95.3}$
& $46.9$ & $\mathbf{53.1}$ & $\mathbf{50.0}$ & $\mathbf{43.8}$ & $\mathbf{34.4}$ \\
\bottomrule
\end{tabular}
}
\end{table}
\begin{table}[t]
\centering
\caption{\textbf{Comparison on Real-World Manipulation.} We report mean values over 32 trials.}
\label{tab:realworld}
\renewcommand{\arraystretch}{1.1}
\resizebox{\linewidth}{!}{%
\begin{tabular}{l @{\hspace{2.5mm}} ccccc @{\hspace{2.5mm}} ccc @{\hspace{2.5mm}} ccccc}
\toprule
\multirow{4}{*}{\textbf{Method}}
& \multicolumn{5}{c}{\texttt{CanSorting}}
& \multicolumn{3}{c}{\texttt{PassTeaPot}}
& \multicolumn{5}{c}{\texttt{ConeCover}} \\
\cmidrule(lr){2-6}\cmidrule(lr){7-9}\cmidrule(lr){10-14}
& \multicolumn{1}{c}{Base} & \multicolumn{2}{c}{$\phi_1$} & \multicolumn{2}{c}{$\phi_2$}
& \multicolumn{1}{c}{Base} & \multicolumn{2}{c}{$\phi_1$}
& \multicolumn{1}{c}{Base} & \multicolumn{2}{c}{$\phi_1$} & \multicolumn{2}{c}{$\phi_2$} \\
\cmidrule(lr){2-2}\cmidrule(lr){3-4}\cmidrule(lr){5-6}\cmidrule(lr){7-7}\cmidrule(lr){8-9}\cmidrule(lr){10-10}\cmidrule(lr){11-12}\cmidrule(lr){13-14}
& \textbf{SR} & \textbf{SR} & \textbf{SRP} & \textbf{SR} & \textbf{SRP}
& \textbf{SR} & \textbf{SR} & \textbf{SRP}
& \textbf{SR} & \textbf{SR} & \textbf{SRP} & \textbf{SR} & \textbf{SRP} \\
\midrule
$\pi_{0.5}$ \cite{black2025pi_} & $\mathbf{100.0}$ & $78.1$ & $43.8$ & $71.9$ & $31.2$
             & $\mathbf{100.0}$ & $81.2$ & $18.8$
             & $43.8$ & $43.8$ & $12.5$ & $34.4$ & $9.4$ \\
\SPES & $93.3$ & $\mathbf{100.0}$ & $\mathbf{100.0}$ & $\mathbf{100.0}$ & $\mathbf{84.4}$
             & $\mathbf{100.0}$ & $\mathbf{100.0}$ & $\mathbf{100.0}$
             & $\mathbf{87.5}$ & $\mathbf{100.0}$ & $\mathbf{100.0}$ & $\mathbf{78.1}$ & $\mathbf{78.1}$ \\
\bottomrule
\end{tabular}
}
\end{table}

\paragraph{Experiment Setup.}
We evaluate across three scenarios: 
(1) \textbf{\textit{Complex Manipulation}} in ManiSkill simulator \cite{mu2021maniskill}: \texttt{StackCube (Fixed)}, \texttt{PegInsert}, \texttt{SeqPushT}.
(2) \textbf{\textit{Long-Horizon TAMP}} in Calvin benchmark \cite{mees2022calvin}: a \texttt{Causal} and an \texttt{Unconstrained} skill set. 
(3) \textbf{\textit{Real-World Manipulation}}: \texttt{PassTeaPot}, \texttt{CanSorting}, and \texttt{ConeCover}. Each task defines a base goal and preference variant(s), across the suite, spanning from object-centric to trajectory-level specifications (\cref{fig:exp}). In (1) and (3), all tested models are trained using the same demonstration dataset and steered via a mechanism suited to their backbone. In (2), we adopt VLA baselines with their best reported weights. Further details on task success criteria and the robot platform are provided in the Appendix.
\paragraph{Metrics.}
We report two orthogonal metrics: (1) \textbf{SR} (\%), the success rate regardless of preference satisfaction; and (2) \textbf{SRP} (\%), the success rate under the specified preference.

\subsection{Task Performance \& Preference Following (Q1)}
\SPES\ achieves the best performance on both SR and SRP across nearly all tasks (\cref{tab:maniskill,tab:realworld}). The baselines exhibit a clear trade-off: reject-filtering methods (Transformer, GMM) enforce preference compliance by discarding non-compliant trajectories, which substantially reduces SR (see $\Delta$= SR-SRP), whereas constraint or prompt injection approaches (DP, OpenVLA, $\pi_{0.5}$) preserve SR but provide weak steering. Moreover, \SPES\ also retains ENAP’s robustness to multimodal demonstrations, as illustrated by the preference-free Base performance on \texttt{SeqPushT}. Nevertheless, steering introduces its own challenge (Equation~\eqref{eq:lowlevel}): the modified $a_t^{\mathrm{prt}}, c_t$ encode the preferred mode, while $\phi_\theta(o_t)$ biases the frozen MLP toward the original learned behavior. This competition most degrades $\phi$-SRP from $\phi$-SR on tasks with complex hybrid dynamics, e.g., \texttt{SeqPushT}.

\begin{table}[t]
\centering
\caption{\textbf{Comparison on Long-Horizon TAMP.} Mean SR for completing the first $k$ skills in the 5-skills ($k/5$), over 8 seeds and 64 trials; Flower$^*$ is Flower fine-tuned on the preference sequence.}
\label{tab:calvin}
\renewcommand{\arraystretch}{1.1}
\resizebox{\linewidth}{!}{%
\begin{tabular}{l c @{\hspace{2.5mm}} cccccc @{\hspace{2.5mm}} cccccc}
\toprule
\multirow{4}{*}{\textbf{Method}} & \multirow{4}{*}{Param (M)}
& \multicolumn{6}{c}{\texttt{Causal}}
& \multicolumn{6}{c}{\texttt{Unconstrained}} \\
\cmidrule(lr){3-8}\cmidrule(lr){9-14}
& & \multicolumn{3}{c}{$\phi_1$} & \multicolumn{3}{c}{$\phi_2$}
  & \multicolumn{3}{c}{$\phi_1$} & \multicolumn{3}{c}{$\phi_2$} \\
\cmidrule(lr){3-5}\cmidrule(lr){6-8}\cmidrule(lr){9-11}\cmidrule(lr){12-14}
& & \textbf{3/5} & \textbf{4/5} & \textbf{5/5} & \textbf{3/5} & \textbf{4/5} & \textbf{5/5}
  & \textbf{3/5} & \textbf{4/5} & \textbf{5/5} & \textbf{3/5} & \textbf{4/5} & \textbf{5/5} \\
\midrule
HULC \cite{hulc}               
& $47$
& $12.5$ & $10.9$ & $1.6$
& $7.8$ & $3.1$ & $0.0$
& $6.3$ & $0.0$ & $0.0$
& $3.1$ & $3.1$ & $3.1$ \\
LCD \cite{lcd}               
& $68$
& $15.6$ & $7.8$ & $7.8$
& $18.8$ & $12.5$ & $12.5$
& $14.1$ & $10.9$ & $3.1$
& $15.6$ & $15.6$ & $9.4$ \\
MDT \cite{mdt}             
& $440$
& $31.3$ & $31.3$ & $3.1$
& $\mathbf{100.0}$ & $\mathbf{100.0}$ & $35.9$
& $14.1$ & $10.9$ & $10.9$
& $15.6$ & $15.6$ & $15.6$ \\
Flower \cite{flower}             
& $947$
& $96.9$ & $57.8$ & $57.8$
& $92.2$ & $42.2$ & $39.1$
& $64.1$ & $51.6$ & $45.3$
& $73.4$ & $70.3$ & $60.9$ \\
Flower$^*$          
& $947$
& $\mathbf{100.0}$ & $87.5$ & $87.5$
& $\mathbf{100.0}$ & $54.7$ & $51.6$
& $78.1$ & $78.1$ & $54.7$
& $93.8$ & $\mathbf{93.8}$ & $76.6$ \\
\midrule
\SPES               
& $571$
& $98.4$ & $\mathbf{93.8}$ & $\mathbf{92.2}$
& $\mathbf{100.0}$ & $81.3$ & $\mathbf{76.6}$
& $\mathbf{90.6}$ & $\mathbf{84.4}$ & $\mathbf{81.3}$
& $\mathbf{100.0}$ & $\mathbf{93.8}$ & $\mathbf{93.8}$ \\
\bottomrule
\end{tabular}
}
\end{table}
\begin{table}[t]
\centering
\caption{\textbf{Ablation Study on Complex Manipulation.} We report mean values over 8 seeds.}
\label{tab:ablation}
\renewcommand{\arraystretch}{1.1}
\resizebox{\linewidth}{!}{%
\begin{tabular}{l @{\hspace{3mm}} ccc @{\hspace{3mm}} ccc @{\hspace{3mm}} ccc}
\toprule
\multirow{4}{*}{\textbf{Method}}
& \multicolumn{3}{c}{\texttt{StackCube (Fixed)}}
& \multicolumn{3}{c}{\texttt{PegInsert}}
& \multicolumn{3}{c}{\texttt{SeqPushT}} \\
\cmidrule(lr){2-4}
\cmidrule(lr){5-7}
\cmidrule(lr){8-10}
& Base \textbf{SR} & $\phi_1$ \textbf{SR} & $\phi_1$\textbf{SRP}
& Base \textbf{SR} & $\phi_1$ \textbf{SR} & $\phi_1$\textbf{SRP}
& Base \textbf{SR} & $\phi_1$ \textbf{SR} & $\phi_1$\textbf{SRP} \\
\midrule
\cellcolor{gray!15}\SPES
& \cellcolor{gray!15}$95.3$ & \cellcolor{gray!15}$\mathbf{98.4}$ & \cellcolor{gray!15}$\mathbf{98.4}$
& \cellcolor{gray!15}$\mathbf{98.4}$ & \cellcolor{gray!15}$\mathbf{95.3}$ & \cellcolor{gray!15}$\mathbf{95.3}$
& \cellcolor{gray!15}$\mathbf{46.9}$ & \cellcolor{gray!15}$\mathbf{53.1}$ & \cellcolor{gray!15}$\mathbf{50.0}$ \\
$\quad \ $ \textit{w\textbackslash o Replay}
& $95.3$ & $35.9$ & $14.1$
& $\mathbf{98.4}$ & $82.8$ & $64.1$
& $\mathbf{46.9}$ & $31.2$ & $15.6$ \\
$\quad \ $ \textit{w\textbackslash o Attention + Clamp}
& $\mathbf{98.4}$ & $37.5$ & $26.6$
& $96.9$ & $93.8$ & $87.5$
& $40.6$ & $34.4$ & $25.0$ \\
\bottomrule
\end{tabular}
}
\end{table}

\begin{figure*}[t]
    \centering
    \includegraphics[width=\linewidth]{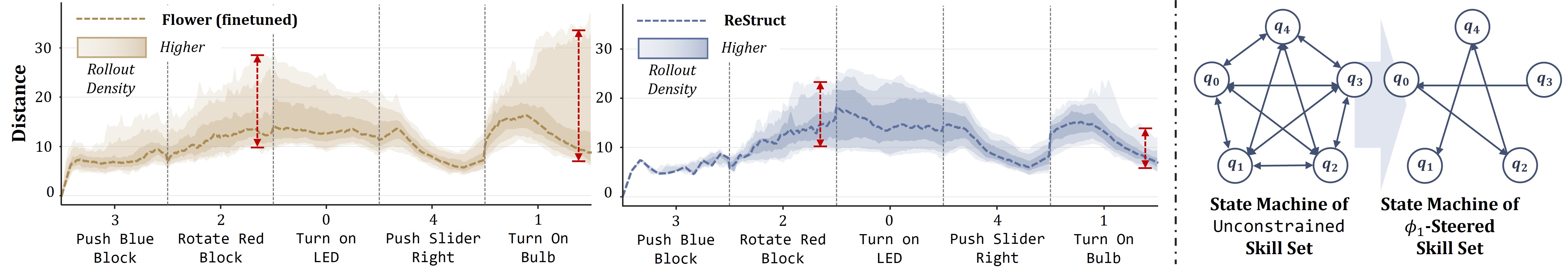}
    \vspace{-12pt}
    \caption{\textbf{Steering a Sequence from a Skill Graph} of \texttt{Unconstrained}-$\phi_1$. (\textbf{Left}) \SPES\ has narrower deviation tubes. (\textbf{Right}) $\phi_1$ steers the fully-connected PMM graph into a sequential chain.}
    \label{fig:error}
\end{figure*}

\subsection{Recovering Ordered Sequences from Unordered Data (Q2)}
As shown in \cref{tab:calvin} (we extract the state machine from the Flower policy), \SPES\ recovers the ordered skill sequence and achieves the highest success rate among all methods. Fine-tuning Flower on a small set of preference-specific trajectories (Flower$^*$) does raise its success rate under the preference; \SPES, by contrast, uses no such trajectories and instead extracts the required sequence directly from the preference-free representation space of the base policy, which raises its own success rate under the preference. In essence, \SPES\ recovers ordered sequences from unordered data. Also, our steered policy is more robust to perturbations: starting from the initial state of a successful demonstration, we execute $64$ noise-perturbed rollouts and measure their per-step deviation from this reference (\cref{fig:error}), and \SPES\ yields a narrower error band than Flower$^*$.


\begin{wrapfigure}{r}{0.44\textwidth}
    \vspace{-5mm}
    \centering
    \includegraphics[width=0.9\linewidth]{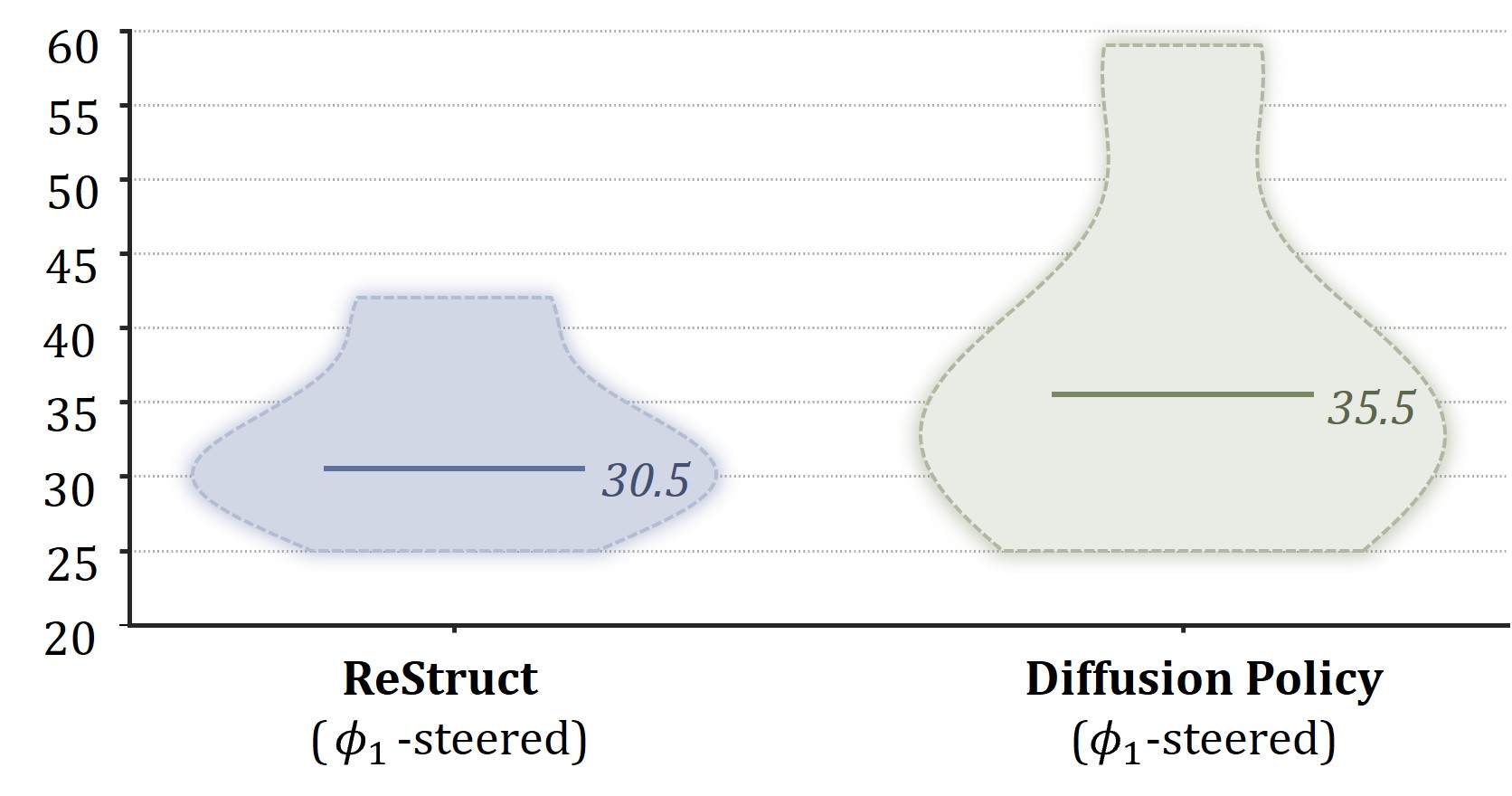}
    \vspace{-8pt}
    \caption{\textbf{Episode-length distribution} over successful $\phi_1$-steered \texttt{PegInsert} rollouts.}
    \label{fig:stateful}
    \vspace{-5pt}
\end{wrapfigure}
\subsection{Decisiveness via Statefulness (Q3)}
\SPES\ inherits \emph{statefulness} from its state-machine-based steering. \emph{First}, the node structure encodes task history as an internal memory. 
\emph{Moreover}, on multimodal tasks such as~\texttt{PegInsert}, \SPES\ can commit to a single preferred mode, thereby reducing runtime mode uncertainty: 
even when both are steered by $\phi_1$, successful \SPES\ episodes are shorter than those of a diffusion policy (\cref{fig:stateful}). 
This reduction in uncertainty can even improve success rates beyond those of the unsteered base policy on multimodal tasks, e.g., the $\phi_1$-SR is higher than Base SR in \texttt{SeqPushT} (\cref{tab:maniskill}).

\begin{table}[t]
\centering
\caption{\textbf{Ablations on Data for PMM Construction and Base Policy Training.} The top row is the full configuration (50 trajectories for PMM construction, 100\% demonstrations for base policy training). (a) reduces the PMM trajectories with the base policy fixed; (b) reduces the base policy training data with the PMM fixed. We report mean values over 8 seeds.}
\label{tab:ablation-data}
\renewcommand{\arraystretch}{1.1}
\resizebox{\linewidth}{!}{%
\begin{tabular}{l @{\hspace{3mm}} ccc @{\hspace{3mm}} ccc @{\hspace{3mm}} ccc}
\toprule
\multirow{2}{*}{\textbf{Method}}
& \multicolumn{3}{c}{\texttt{StackCube (Fixed)}}
& \multicolumn{3}{c}{\texttt{PegInsert}}
& \multicolumn{3}{c}{\texttt{SeqPushT}} \\
\cmidrule(lr){2-4}
\cmidrule(lr){5-7}
\cmidrule(lr){8-10}
& Base \textbf{SR} & $\phi_1$ \textbf{SR} & $\phi_1$\textbf{SRP}
& Base \textbf{SR} & $\phi_1$ \textbf{SR} & $\phi_1$\textbf{SRP}
& Base \textbf{SR} & $\phi_1$ \textbf{SR} & $\phi_1$\textbf{SRP} \\
\midrule
\rowcolor{gray!15}
\SPES\ (50 Trajs, 100\%)
& $\mathbf{95.3}$ & $\mathbf{98.4}$ & $\mathbf{98.4}$
& $\mathbf{98.4}$ & $\mathbf{95.3}$ & $\mathbf{95.3}$
& $\mathbf{46.9}$ & $\mathbf{53.1}$ & $\mathbf{50.0}$ \\
\midrule
\rowcolor{gray!8}
\multicolumn{10}{l}{\textit{- (a) PMM Construction Trajectories (Base Policy Fixed at 100\%)}} \\
$\quad \ $ \textit{- 30 Trajs}
& $93.8$ & $90.6$ & $87.5$
& $90.6$ & $89.1$ & $89.1$
& $37.5$ & $42.2$ & $35.9$ \\
$\quad \ $ \textit{- 10 Trajs}
& $90.6$ & $84.4$ & $79.7$
& $92.2$ & $84.4$ & $81.3$
& $34.4$ & $32.8$ & $25.0$ \\
\rowcolor{gray!8}
\multicolumn{10}{l}{\textit{- (b) Base Policy Training Data (PMM Fixed at 50 Trajs)}} \\
$\quad \ $ \textit{- 75\% Trajs}
& $93.8$ & $93.8$ & $90.6$
& $90.6$ & $93.8$ & $93.8$
& $37.5$ & $45.3$ & $40.6$ \\
$\quad \ $ \textit{- 50\% Trajs}
& $84.4$ & $87.5$ & $84.4$
& $89.1$ & $85.9$ & $82.8$
& $31.3$ & $35.9$ & $34.4$ \\
\bottomrule
\end{tabular}
}
\end{table}

\subsection{Ablation Studies.}
We first show the two components of \SPES's action prior (\cref{tab:ablation}). \textit{w\textbackslash o Replay} skips the replay-based re-grounding, and \textit{w\textbackslash o Attention+Clamp} collapses each list and its attention into a single fixed mean action. Each removal hurts task performance and preference adherence, which confirms the validity of our designs.

Table~\ref{tab:ablation-data} further exposes a clean functional decoupling. Reducing the trajectories used to construct the PMM primarily harms preference satisfaction (SRP) while leaving raw task completion (SR) only mildly affected; conversely, undertraining the base policy primarily harms SR while leaving SRP relatively stable, since the symbolic structure that carries the preference handle remains intact. This asymmetry holds across all three \texttt{Complex Manipulation} tasks, including the more dynamics-heavy \texttt{SeqPushT}. The results reveal that preference adherence depends primarily on whether the PMM carries sufficiently rich priors, while overall task performance hinges on the scale of data and training devoted to the base policy.

\section{Limitation}
\emph{First}, \SPES\ builds on ENAP, and extending it to other policy classes remains open, since distilling policies into ENAP incurs a performance gap---future direction is reducing the gap. 
\emph{Second}, both task success and preference compliance depend on the VLM-generated LTL formula. This may be alleviated through iterative, multi-round formula refinement. 
\emph{Third}, our steerable scope (\cref{rem:scope}) is bounded: scaling to richer OOD regimes and eliciting new skills remain unexplored.


\section{Conclusion.}
We propose \SPES, a zero-shot behavior steering framework that satisfies preferences at inference time by reconfiguring a policy’s task structure without retraining. Specifically, we first transform the preference into an automaton and generate a steered task structure based on it, while keeping the base policy network frozen. An action prior recomputation process is then used to bridge the gap between the modified high-level structure and the original controller. Extensive experiments validate \SPES\ consistently leads in both success and preference compliance. More broadly, \SPES\ provides a viable path for everyday users to reshape their robots by language, not retraining.

\newpage

\bibliography{example}  
\newpage

\appendix
\section{Related Works}
\subsection{Preference Specification for Robot Policies}
\label{app:rw-preference}
Deploying a learned policy to real users almost always reveals a gap between what it was trained to do and what each user actually wants. In the domain, preference specification spans a broad spectrum. \textit{Scalar comparisons} remain canonical: feedback can train rewards \cite{r1,r2,pan2026wonder}, and DPO-style updates can align diffusion policies from pairwise preferences without an explicit reward \cite{r3}. \textit{Natural language} provides a more flexible substrate, from describing deviations in the source demonstrations \cite{r4} to grounding open-vocabulary object goals \cite{r4,r5,r6,r7}. 
\textit{Demonstrations} also encode intent implicitly, from hand-held grippers \cite{r8} and web videos \cite{r9} to egocentric smart-glass recordings \cite{r10} that scale data collection toward zero-shot deployment. \textit{Visual goals} reduce linguistic ambiguity through image-edited subgoals \cite{r11} and intentionally sparse sketches \cite{r12}, while \textit{formal specifications} encode intent as automata or temporal-logic formulas, increasingly coupled with LLM-based language-to-spec translation \cite{r13,r14}.

Policy-side preference integration can be grouped into three regimes. \textit{Training-time methods} re-optimize or fine-tune policies under the preference signal \cite{r1,r2,r3,r8,r9,r10}; \textit{prompt-level methods} condition language-aware VLAs \cite{r5} or pass LLM-generated plans to low-level controllers \cite{r7}; and \textit{neuro-symbolic methods} route preferences through symbolic intermediaries layered over the policy, such as LLM-scored affordances over learned skill libraries \cite{r15}, executable code calling neural primitives \cite{r16}, or language-translated formal specifications consumed by planners \cite{r14}.

These approaches expose two structural limitations. Controller-alignment methods offer broad preference expressivity, but require re-optimizing or fine-tuning the policy for each new user preference \cite{r1,r3,r8,r9,r10}, making them impractical after deployment. Symbolic-routing methods avoid modifying the controller by inserting a symbolic scaffold \cite{r7,r13,r14,r15,r16}, but this scaffold is typically defined by a fixed designer-specified vocabulary, with a fixed symbol-to-controller interface that prevents symbolic edits from reliably propagating to the executed behavior. What is needed, then, is a steering route that avoids retraining yet still propagates the user's preference all the way into the controller's behavior---which is precisely what \SPES\ provides.

\subsection{Neuro-symbolic Policies for Manipulation}
\label{app:rw-nesy}
Neuro-symbolic methods in manipulation aim to combine the compositional structure and interpretability of symbolic reasoning with the perceptual and sensorimotor flexibility of neural networks. Classical task and motion planning frameworks such as PDDLStream \cite{s1} established a clean factoring---discrete symbolic search over manually specified predicates paired with continuous motion planners---but their reliance on hand-authored symbolic vocabularies sharply limited adaptability. Subsequent work has therefore pushed each layer of this stack toward learnable substitutes. In the action layer, hierarchical and option-based methods replace fixed primitives with discoverable skills \cite{s2}, later evolving into great parameterized skills \cite{s3,s4} that admit continuous control within otherwise symbolic plans. In the state layer, predicate-learning approaches ground continuous sensorimotor signals into discrete symbols \cite{s5,s6}---through neuro-symbolic concept learners \cite{s8}, or self-supervised automaton extraction directly from visuomotor data \cite{pan2026emergent}. In the model layer, the symbolic transition dynamics themselves are inferred, from probabilistic rules in stochastic domains \cite{s10} to graph-neural relational models \cite{s11} and hybrid automata for stochastic transitions \cite{s12}. A more recent and complementary line offloads the high-level structure to pretrained foundation models \cite{r15,r16,r14}.

While these approaches successfully marry learning with symbolic structure, they share a common architectural commitment: the symbolic layer is \emph{external} to the controller---either hand-authored, learned as a separate module, or generated by a foundation model on top of a frozen low-level policy---and the controller acts as a black box beneath it. \SPES{} departs from this pattern by re-computing the policy's symbolic-level prior under each user preference and propagating it back into the low-level controller, so that the preference is realized as a physically-aware steering signal rather than a detached symbolic plan layered above the policy.

\subsection{Inference-time Adaptation of Learned Policies}
\label{app:rw-inferenceadapt}
A recent line of work asks how to bias a deployed generative policy without retraining it \cite{pan2025planning,xu2026dream}. Most methods operate inside the diffusion denoising loop: human sketches and clicks bias the action sampling \cite{wang2025inference}, an external dynamics model guides toward visual outcomes \cite{du2026dynaguide}, or Feynman-Kac particles resample intermediate trajectories under arbitrary rewards \cite{singhal2025general}. Decoding-level methods instead choose among policy proposals---bidirectional decoding searches for actions consistent across timesteps \cite{z4} and contrastive decoding refocuses attention on task-relevant objects \cite{z5}---without altering policy weights. Latent-space approaches run reinforcement learning over the diffusion policy's noise input to adapt it online \cite{z6}. For VLA-class generalists, pretrained VLMs act as runtime verifiers, selecting action plans whose predicted outcomes align with the VLA's textual reasoning \cite{z7} or with downstream task goals \cite{z8}.

These methods steer within the action distribution, which restricts the preferences they can handle to those that are approximately differentiable, such as a gradient signal or a sampleable verifier. Genuine steering, however, requires reaching beyond the action distribution: forming a semantic understanding of the real-world complex preference, and feeding that understanding back into the action distribution itself. \SPES\ provides this route via computing a symbolic-level interpretation of the user preference and propagating it back through the controller's action distribution.

\section{Experimental Setups}
\subsection{Embodiments and Training Compute}
\textbf{Complex Manipulation} and \textbf{Long-Horizon TAMP} both run on a Franka Emika Panda, with average demonstration lengths of $\sim$13K and $\sim$30K steps, respectively; the \textbf{Real-World} suite uses a Kinova Gen3 in a noisy, data-sparse setting with much shorter $\sim$1.6K-step demonstrations. All experiments are trained on a single NVIDIA RTX 4090.

\subsection{Task Definition}
\paragraph{1) \texttt{StackCube (Fixed)}.}
The agent must stack two colored cubes (Red/Green) in any valid vertical configuration. The initial positions of the two cubes are fixed, located on opposite sides of the arm. A trial is considered successful only if the top cube aligns within half-width of the bottom cube, remains static, and is fully released by the gripper.

\paragraph{2) \texttt{PegInsert}.} 
This task requires picking up a bi-colored peg and inserting either the orange or white end into a randomly positioned hole on the box. The environment features significant variability: peg lengths are sampled from $U[0.15, 0.20]\text{m}$ and radii from $U[0.015, 0.025]\text{m}$, with initial rotation poses randomized. Success is defined as inserting either end at least $0.03\text{m}$ into the hole (radius $r_{\text{peg}}+0.02\text{m}$).

\paragraph{3) \texttt{SeqPushT}.}
A variant with sequential goals of the Push-T task. The robot must maneuver a T-shaped block to two potential target poses in any order. The T-block's initial pose is fully randomized within a $0.076m \times 0.20m$ respawn box. Success is achieved if the block reaches the two target poses in sequence, with its Intersection-over-Union (IoU) exceeding $0.90$ at each target pose.

\paragraph{4) \texttt{Unconstrained}.}
The agent completes five atomic skills---\texttt{turn\_on\_led} (LED on), \texttt{turn\_on\_bulb} (bulb on), \texttt{rotate\_block} (red block rotated), \texttt{push\_left} (blue block pushed left), and \texttt{move\_slider} (slider moved right)---in the order specified by our preference. These skills are mutually independent; success requires all five in the prescribed order.

\paragraph{5) \texttt{Causal}.}
The agent completes five atomic skills---\texttt{rotate\_block} (blue block rotated), \texttt{close\_drawer} (drawer closed), \texttt{open\_drawer} (drawer opened), \texttt{push\_right} (pink block pushed right), and \texttt{push\_left} (blue block pushed left)---in the order specified by our preference. \texttt{close\_drawer} causally depends on \texttt{open\_drawer}, constraining the order of skill attempts.

\paragraph{6) \texttt{CanSorting}.}
It involves sorting three colored cans (red, green, blue) into the same-colored bowls fixed on the table. Our preference, however, asks the agent to instead place each can into a bowl of a target color independent of its own---an OOD behavior demanding compositional generalization.

\paragraph{7) \texttt{PassTeaPot}.}
The arm grasps a teapot's handle at a fixed pose and carries it to the table's far end, with a power strip placed randomly along one side. Demonstrations cover trajectories both near and far from the strip; our preference asks for look-ahead trajectory shaping.

\paragraph{8) \texttt{ConeCover}.}
The arm grasps a cone from a fixed source and covers two fixed differently-colored blocks (blue, green) in any order. Cones are visually identical, requiring spatial memory---for example, lifting the cone already on the blue block and redirecting it onto the
green one.

\begin{figure*}
    \centering
    \includegraphics[width=\linewidth]{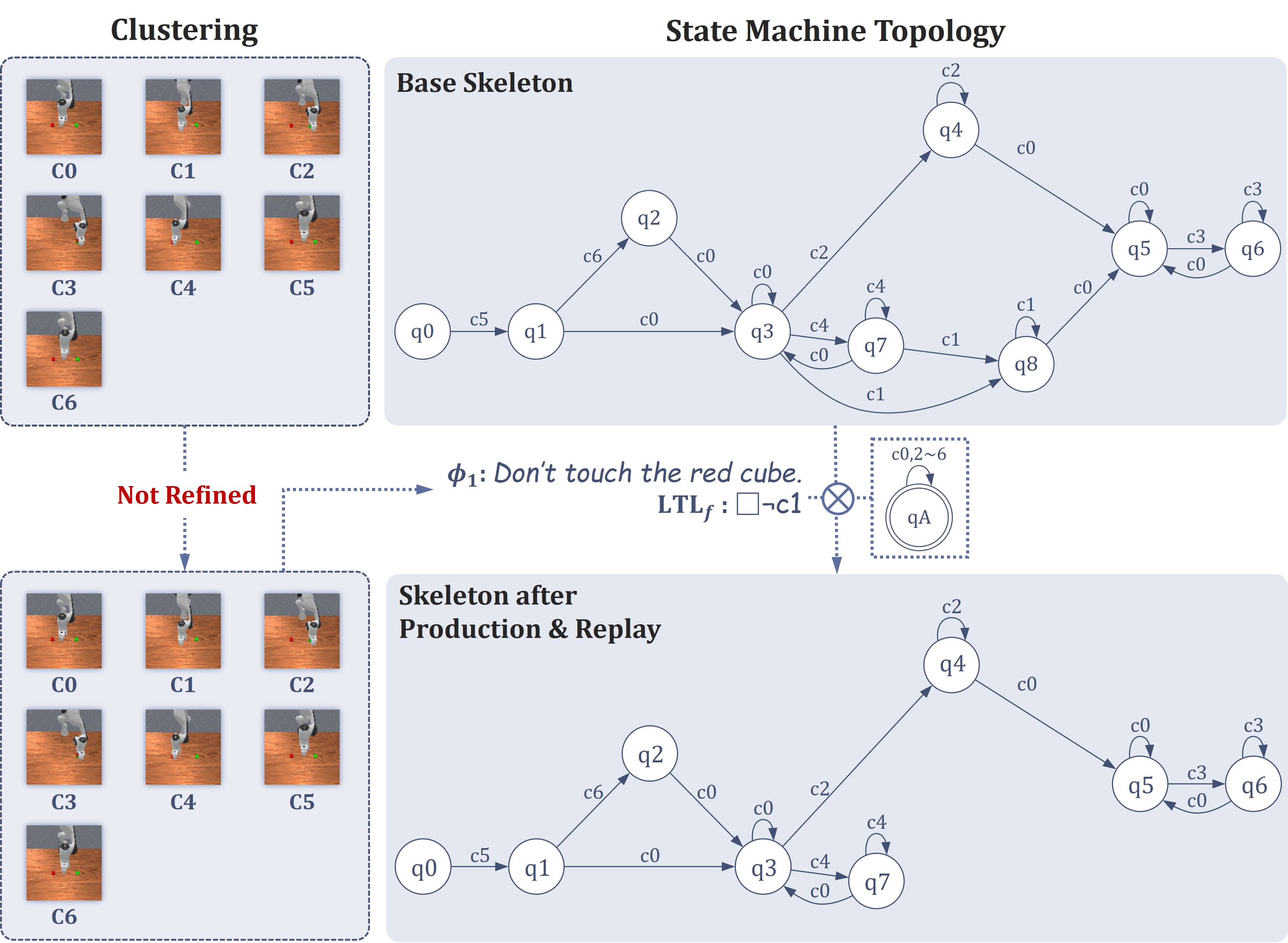}
    \vspace{-12pt}
    \caption{
    \textbf{Diagram of Skeleton Variations for \texttt{StackCube (Fixed)}.} The base skeleton and the DFA of preference $\phi_1$ are synthesized into a steered state machine, with an unrefined cluster table.
    }
    \label{fig:stack_pmm}
    \vspace{-0pt}
\end{figure*}

\begin{figure*}
    \centering
    \includegraphics[width=\linewidth]{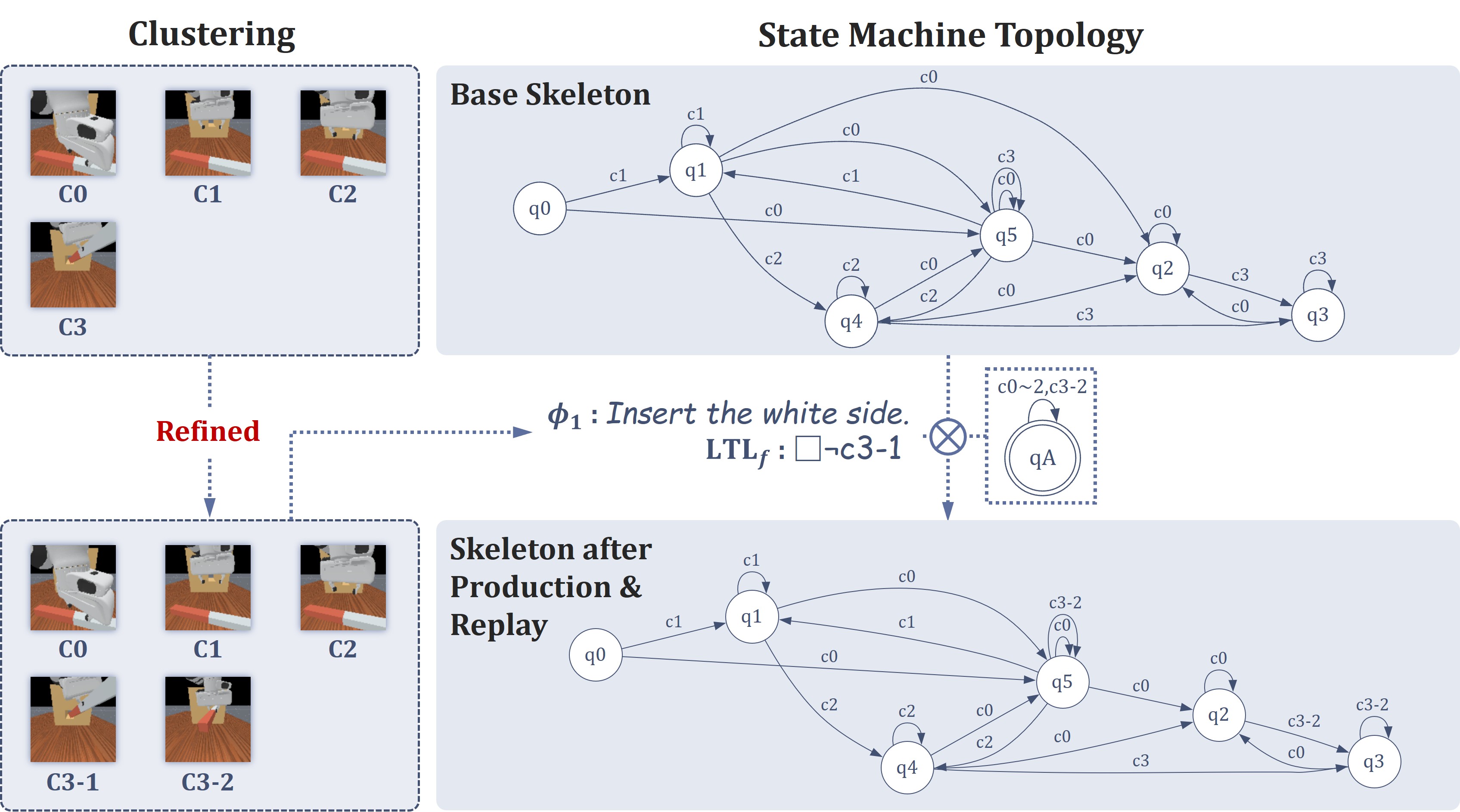}
    \vspace{-12pt}
    \caption{
    \textbf{Diagram of Skeleton Variations for \texttt{PegInsert}.} The base skeleton and the DFA of preference $\phi_1$ are synthesized into a steered state machine, with a refined cluster table.
    }
    \label{fig:peg_pmm}
    \vspace{-0pt}
\end{figure*}

\begin{figure*}
    \centering
    \includegraphics[width=\linewidth]{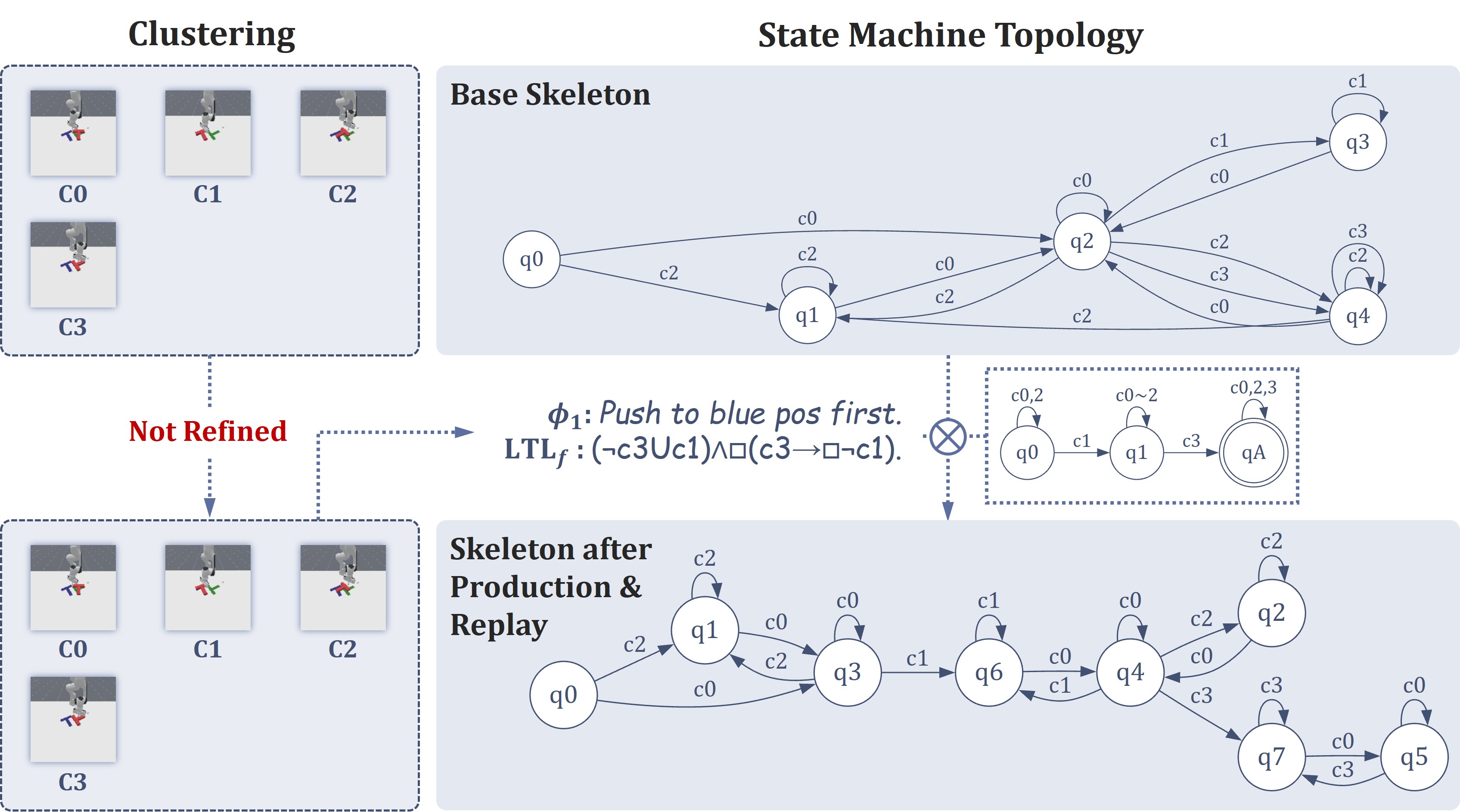}
    \vspace{-12pt}
    \caption{
    \textbf{Diagram of Skeleton Variations for \texttt{SeqPushT}.} The base skeleton and the DFA of preference $\phi_1$ are synthesized into a steered state machine, with an unrefined cluster table.
    }
    \label{fig:seq_pmm}
    \vspace{-0pt}
\end{figure*}

\begin{figure*}
    \centering
    \includegraphics[width=\linewidth]{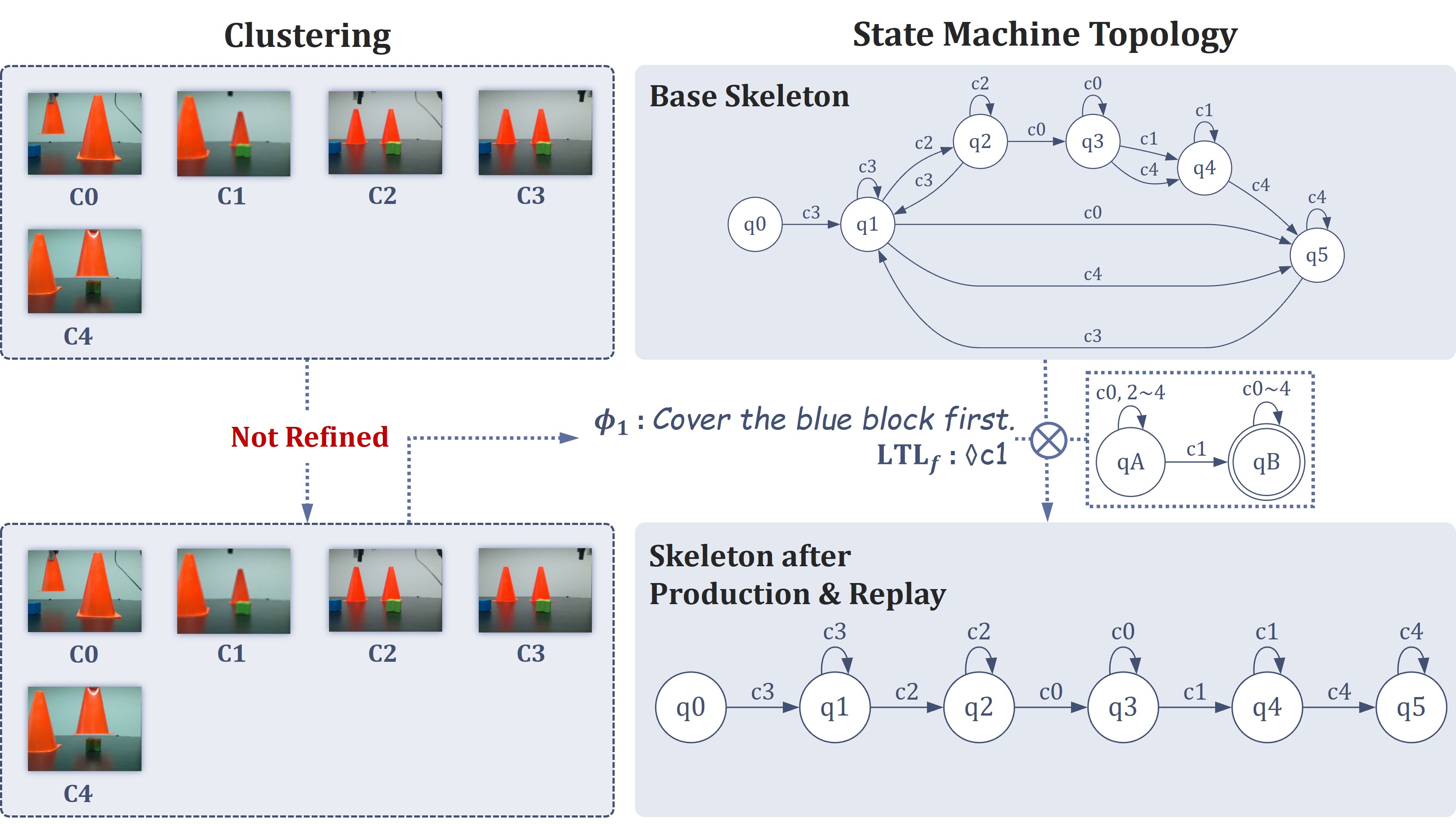}
    \vspace{-12pt}
    \caption{
    \textbf{Diagram of Skeleton Variations for \texttt{ConeCover}.} The base skeleton and the DFA of preference $\phi_1$ are synthesized into a steered state machine, with an unrefined cluster table.
    }
    \label{fig:cone_pmm}
    \vspace{-0pt}
\end{figure*}

\begin{figure*}
    \centering
    \includegraphics[width=\linewidth]{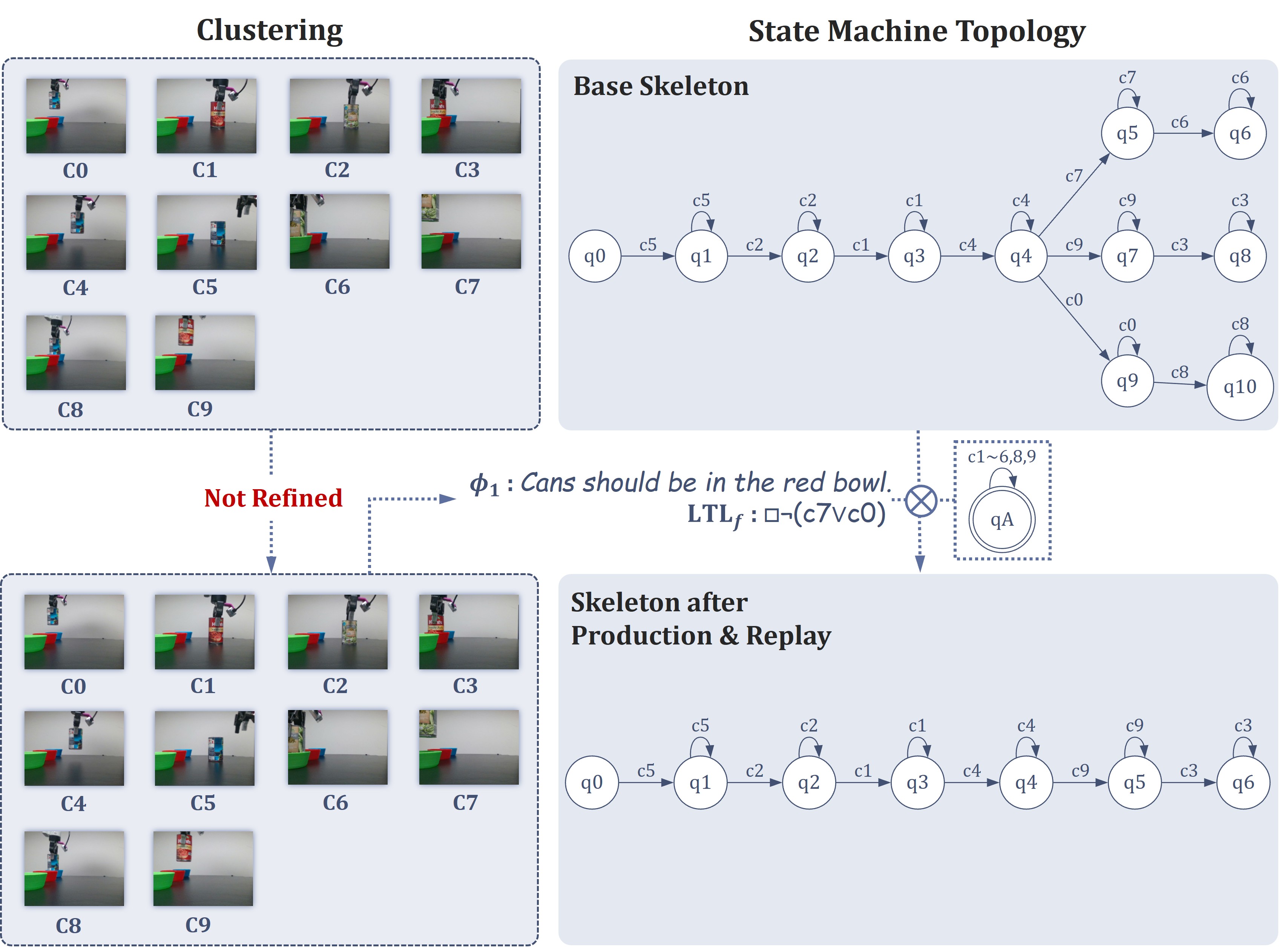}
    \vspace{-12pt}
    \caption{
    \textbf{Diagram of Skeleton Variations for \texttt{CanSorting}.} The base skeleton and the DFA of preference $\phi_1$ are synthesized into a steered state machine, with an unrefined cluster table.
    }
    \label{fig:can_pmm}
    \vspace{-0pt}
\end{figure*}

\begin{figure*}
    \centering
    \includegraphics[width=\linewidth]{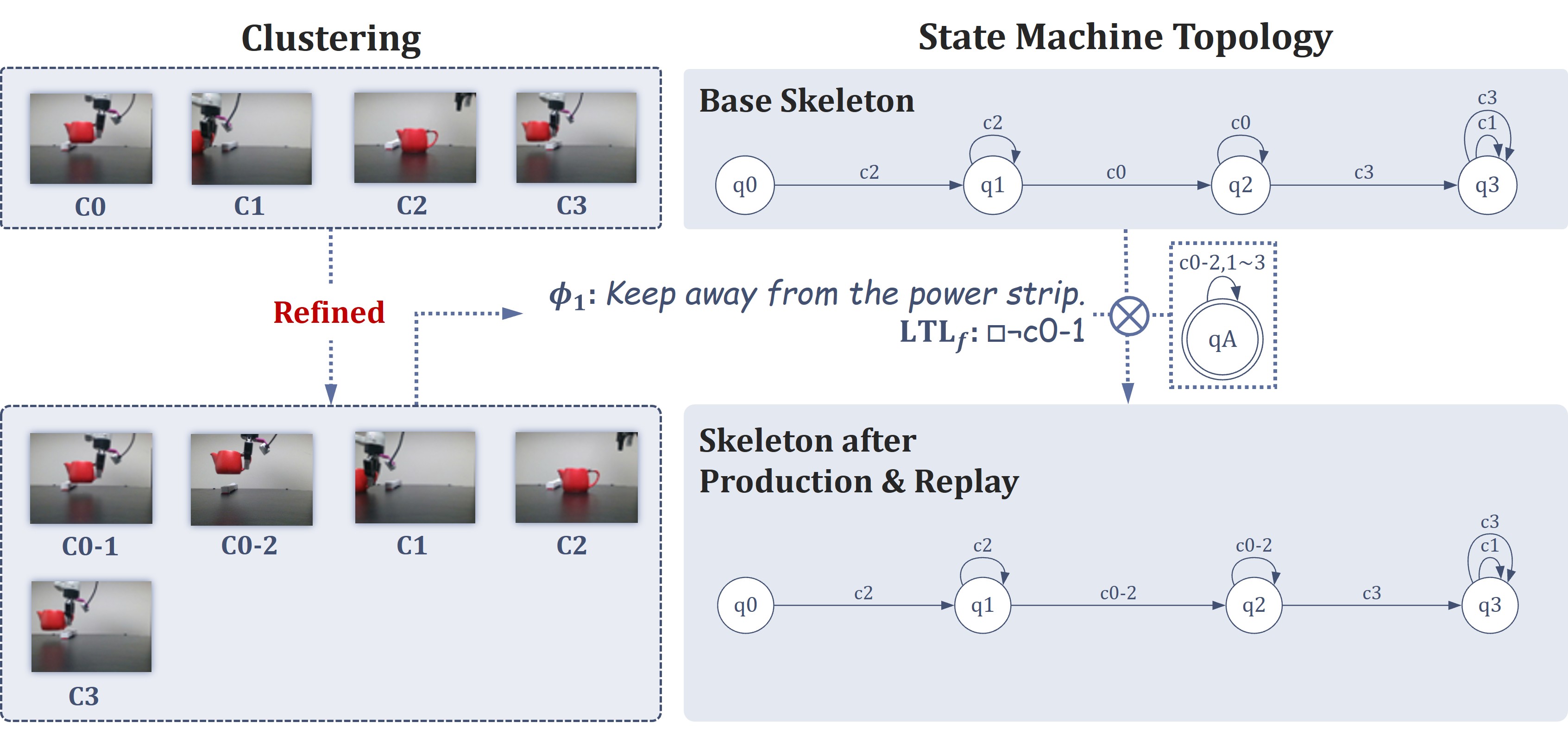}
    \vspace{-12pt}
    \caption{
    \textbf{Diagram of Skeleton Variations for \texttt{PassTeaPot}.} The base skeleton and the DFA of preference $\phi_1$ are synthesized into a steered state machine, with a refined cluster table.
    }
    \label{fig:pass_pmm}
    \vspace{-0pt}
\end{figure*}

\section{Walkthroughs of Product PMM Construction}
We illustrate the skeleton refinement pipeline on the three \textbf{Complex Manipulation} tasks in \cref{fig:stack_pmm}, \cref{fig:peg_pmm}, and \cref{fig:seq_pmm} and three \textbf{Real-World} tasks in \cref{fig:cone_pmm}, \cref{fig:can_pmm},  and \cref{fig:pass_pmm}. Each figure pairs the cluster table (refined or not) with the state-machine topology, showing how preference $\phi_1$ compiles
into a DFA and product-composes with the base skeleton to yield the steered state machine. For readability, the nodes in the \emph{Skeleton after Production \& Replay} are labeled using the same symbols as the base skeleton, such as $q_0$. However, these nodes actually represent product states and should not be interpreted as the original base nodes. Moreover, this post-replay topology already reflects the pruning of edges that are not traversed by accepted trajectories, so the action lists carried by the surviving edges have likewise been re-populated.


\section{VLM Prompt Templates}
Two VLM queries operationalize the preference compilation pipeline: \textit{First}, the \emph{Cluster Refinement Query} ingests the task description, the $M$ learned clusters (each summarized by $N$ representative images), a small set of demonstration trajectories, and the user preference; it decides whether one cluster $C_i$ should be partitioned into $K$ sub-clusters, returning either NO or YES together with an explicit image-to-sub-cluster assignment that sums to $N$. \textit{Second}, the \emph{LTL Translation Query} then runs on the refined clusters and emits a single LTL formula over the cluster alphabet $c_0, \ldots, c_{M-1}$ in a fixed underscore-and-parenthesis syntax that the downstream DFA converter consumes verbatim.

\vspace{5pt}

\begin{tcolorbox}[
  colback=gray!5,
  colframe=gray!30,
  coltitle=black,
  title={Prompt 1: Cluster Refinement Query},
  fonttitle=\bfseries,
  fontupper=\ttfamily\small,
  label={box:cluster-refinement},
  breakable
]
You are an analyst examining state clusters learned from robot manipulation demonstrations. Your job is to decide whether a single cluster needs to be split into sub-clusters so that a user preference can be enforced over the resulting skeleton.
\\ \\
\# Background \\
- A state-machine skeleton has been learned from visuomotor demonstrations. Each cluster $C_i$ groups visually and behaviorally similar states; transitions between clusters form the skeleton. \\
- A cluster requires refinement when the user preference would treat two subsets of that cluster differently---for example, some states inside $C_i$ satisfy the preference while others violate it.
\\ \\
\# Inputs \\
\#\# Task \\
\{\{TASK\_DESCRIPTION\}\} \\
\#\# Cluster representatives \\
There are $M$ = \{\{NUM\_CLUSTERS\}\} clusters. For each cluster $C_i$, you receive $N$ = \{\{N\_IMAGES\_PER\_CLUSTER\}\} representative images, indexed $1$ through $N$:
\[
\rm [IMAGES: C\_0]   [IMAGES: C\_1]   ...   [IMAGES: C\_\{M-1\}]
\]

\#\# Representative trajectories \\
Each trajectory is a sequence of (cluster\_id, action) pairs sampled from a demonstration, formatted as:
\[
  \text{Trajectory } k: (c_0, a_0) \to (c_1, a_1) \to \cdots \to (c_T, a_T)
\]

\{\{TRAJECTORIES\}\}\\
\#\# User preference\\
\{\{USER\_PREFERENCE\}\}
\\ \\
\# Rules \\
- Identify AT MOST ONE cluster $C_i$ to split in this response. If multiple clusters look refinable, pick the one most directly implicated by the preference.

- The split must PARTITION the $N$ representative images of $C_i$ into $K$ sub-clusters $C_i$-1, $C_i$-2, $\ldots$, $C_i$-$K$ (with $K$$\geq$$2$). Each image is assigned to exactly one sub-cluster, and the sub-cluster sizes sum to $N$.

- Use the cluster images AND the trajectories jointly when judging whether a sub-distinction exists. Trajectories tell you how a cluster is used over time; the images tell you what states inside it look like.
\\ \\
\# Output format \\
Respond in EXACTLY one of the two formats below; do not include any prose, reasoning, or text outside these formats.

[Case 1] No cluster needs splitting: \\
NO

[Case 2] Cluster $C_i$ should be split into $K$ sub-clusters: \\
YES \\
Cluster: C\_i \\
Number of sub-clusters: K \\
Assignment (image indices in $[1, N]$; each index appears in exactly one sub-cluster): \\
- C\_i-1: images <indices>, criterion: <one-sentence visual/semantic> \\
- C\_i-2: images <indices>, criterion: <one-sentence visual/semantic> \\
- \ldots \\
- C\_i-K: images <indices>, criterion: <one-sentence visual/semantic> \\
\end{tcolorbox}

\vspace{5pt}

\begin{tcolorbox}[
  colback=gray!5,
  colframe=gray!30,
  coltitle=black,
  title={Prompt 2: LTL Translation Query},
  fonttitle=\bfseries,
  fontupper=\ttfamily\small,
  label={box:ltl-translation},
  breakable
]
You are an analyst translating a user preference about a robot
manipulation task into a Linear Temporal Logic (LTL) formula over
the task's learned cluster alphabet.

\

\# Background

A state-machine skeleton has been learned from demonstrations; each state corresponds to a refined cluster $C_i$. To enforce the preference at runtime, we compile it into an LTL formula whose atomic propositions are cluster identifiers ($c_0, c_1, \ldots$). The formula is then converted to a DFA and product-composed with the skeleton.

\

\# Inputs

\#\# Task

\{\{TASK\_DESCRIPTION\}\}

\#\# Refined clusters

There are $M$ = \{\{NUM\_CLUSTERS\}\} refined clusters. For each cluster
$C_i$ you receive \{\{N\_IMAGES\_PER\_CLUSTER\}\} representative images:
\[
\rm [IMAGES: C\_0]   [IMAGES: C\_1]   ...   [IMAGES: C\_\{M-1\}]
\]

\#\# User preference

\{\{USER\_PREFERENCE\}\}

\

\# Procedure

- For each cluster $C_i$, infer its semantic role from the images and the task description (e.g., "pick\_teapot", "place\_in\_red\_bowl", "approach\_power\_strip"). These names are for your own reasoning; they will NOT appear in the output.

- Map the user preference onto cluster identifiers---identify which clusters the preference constrains, and the temporal relation among them (avoid, require, sequence, until, ...).

- Express the preference as an LTL formula whose atomic propositions are the cluster identifiers $c_i$.

\

\# LTL syntax (use EXACTLY this notation)

Atomic propositions: c0, c1, ..., c\{M-1\}

Operators (connect via underscores; group with parentheses):

- "not\_X"        negation of X

- "G\_X"          X holds globally (always)

- "F\_X"          X holds eventually

- "X\_U\_Y"        X holds until Y

- "X\_to\_Y"       X implies Y

- "X\_and\_Y"      conjunction

- "X\_or\_Y"       disjunction

Examples:

- "never enter c3" -> "G\_not\_c3"

- "eventually reach c5" -> "F\_c5"

- "avoid c1 until c3, then if c1 occurs never reach c3": -> "(not\_c1\_U\_c3)\_and\_G(c1\_to\_G(not\_c3))"

\

\# Output format

Respond with EXACTLY ONE LINE: the LTL formula in the syntax above. Do not include any prose, reasoning, justification, code fences, or markdown.
\end{tcolorbox}


\end{document}